\documentclass{article}

\PassOptionsToPackage{numbers, sort&compress}{natbib}

\usepackage[preprint]{neurips_2026}
\usepackage{microtype}
\usepackage{enumitem}
\usepackage{graphicx}
\usepackage{subcaption}
\usepackage{booktabs}
\usepackage{kotex}
\usepackage{multirow}
\usepackage{xcolor}
\usepackage{pifont} 
\usepackage[table]{xcolor}
\usepackage{hyperref}

\usepackage{amsmath}
\usepackage{amssymb}
\usepackage{mathtools}
\usepackage{amsthm}
\usepackage{wrapfig}
\usepackage[capitalize,noabbrev]{cleveref}

\newcommand{\blackx}{\ding{55}}
\newcommand{\blackv}{\ding{51}}
\definecolor{hlgreen}{RGB}{220,245,220}   
\definecolor{hlblue}{RGB}{220,235,250}    
\definecolor{hlpurple}{RGB}{235,225,245}  
\definecolor{hlred}{RGB}{255,230,230}  
\newcommand{\graybox}[1]{%
  \raisebox{0pt}{\smash{\colorbox{gray!10}{#1}}}%
}
\newcommand{\ours}{CSFM}

\usepackage[utf8]{inputenc} 
\usepackage[T1]{fontenc}    
\usepackage{hyperref}       
\usepackage{url}            
\usepackage{booktabs}       
\usepackage{amsfonts}       
\usepackage{nicefrac}       
\usepackage{microtype}      
\usepackage{xcolor}         

\title{Better Source, Better Flow: \\
Learning Condition-Dependent Source Distribution for Flow Matching}

\author{
\begin{tabular}{c}
Junwan Kim\textsuperscript{1}\thanks{Equal contribution.} \quad
Jiho Park\textsuperscript{2}\footnotemark[1] \quad
Seonghu Jeon\textsuperscript{2} \quad
Seungryong Kim\textsuperscript{2}\thanks{Corresponding author}
\\[1ex]
{\normalfont \textsuperscript{1}New York University \quad \textsuperscript{2}KAIST AI}
\\[1ex]
{\normalfont \href{https://junwankimm.github.io/CSFM/}
{\textcolor{purple}{\tt https://junwankimm.github.io/CSFM/}}}
\end{tabular}
}

\begin{document}

\maketitle

\begin{figure*}[h]
    \centering
    \vspace{-1.5em}
\includegraphics[width=1.0\linewidth]
    {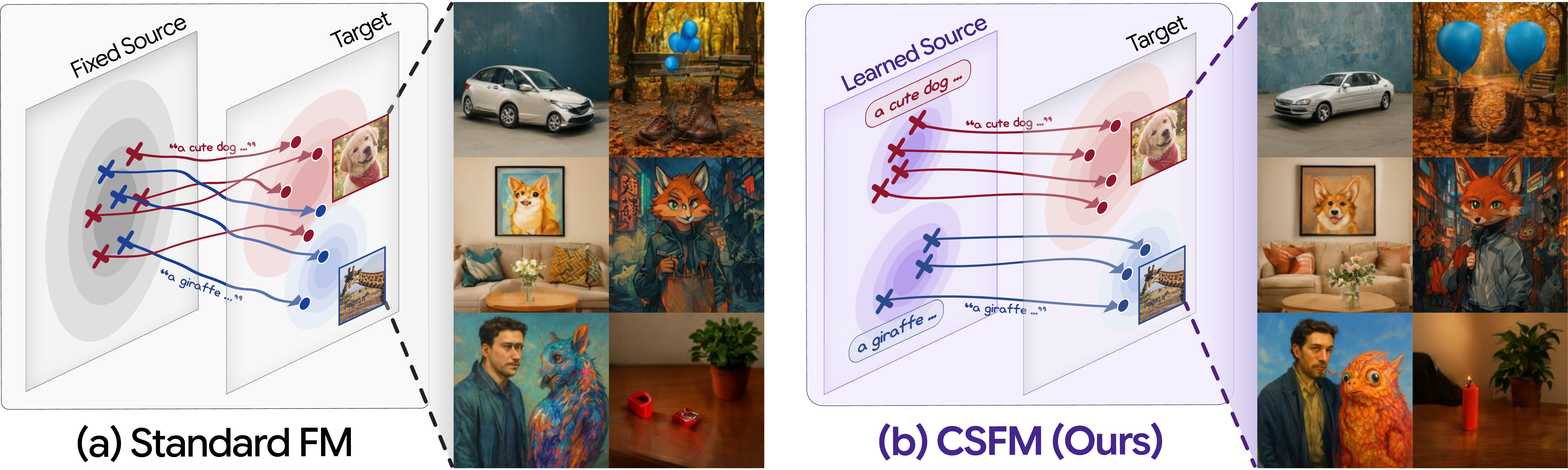}
    \caption{\textbf{Condition-dependent Source Flow Matching (CSFM). } Flow matching does not require the source distribution to be a fixed standard Gaussian.   
    We leverage this flexibility by learning a condition-dependent source distribution, making flow matching easier to train and improving generation performance. Qualitative examples illustrate improved generation quality.
    }
    \label{fig:main_1column}\vspace{-20pt}
    \vspace{1em}
\end{figure*}

\begin{abstract}
\label{abstract}
Flow matching has recently emerged as a promising alternative to
diffusion-based generative models, particularly for text-to-image
generation. Although flow matching places no restriction on the source
distribution, most existing systems still inherit a standard Gaussian
from diffusion models, and the source is rarely treated as an
optimization target at this scale. Recent works have begun to revisit
this choice through condition-dependent or learned sources, yet
evidence that such designs are effective in modern text-to-image
systems---with high-dimensional latents and tightly integrated
conditioning---remains limited. In this work, we study
condition-dependent source distributions for flow matching along three
axes: \emph{why} source learning helps, through the lens of the
intrinsic variance term in the flow-matching objective; \emph{how} to
make it work in modern text-to-image systems, where variance-only
regularization and directional source--target alignment are critical
for stable end-to-end training; and \emph{when} it is most beneficial,
by connecting source design to recent representational advances in
generative modeling and identifying target representation regimes in
which learning the source yields the largest gains. Extensive
experiments across multiple text-to-image benchmarks, architectures, and
scales demonstrate that principled source design yields consistent and
robust improvements---including up to $\mathbf{3.01\times}$ faster
convergence in FID and $\mathbf{2.48\times}$ in CLIP Score---and
outperforms representative condition-dependent source and condition-aware coupling methods.

\end{abstract}

\section{Introduction}
\label{sec:intro}
Flow Matching~\cite{lipman2022fm, liu2022rf} has recently emerged as a powerful framework for generative modeling, demonstrating strong performance across many domains. In particular, text-to-image (T2I) systems show that \textit{conditional} flow matching\footnotemark can generate high-fidelity images with strong prompt adherence~\cite{esser2024sd3, flux2024, mao2026wan, qin2025lumina}, suggesting that it is a promising alternative to diffusion-based models for conditional image generation.

\footnotetext{\hypertarget{cfm}{In} this paper, \emph{conditional} flow matching refers to flow matching conditioned on an external variable. This should not be confused with Conditional Flow Matching (CFM)~\cite{lipman2022fm}, where ``conditional'' instead refers to conditioning on $X_1$.}

At its core, flow matching learns a continuous-time velocity field that transports samples from a \textit{source} distribution to a \textit{target} distribution. Unlike diffusion models, which rely on stochastic noise injection~\cite{ho2020ddpm, song2020ddim} or denoising score matching~\cite{song2020score, song2019ncsn}, flow matching directly models deterministic dynamics. More importantly, it places no restriction on the source distribution. This flexibility is especially appealing for conditional generation, since the conditioning signal can be used not only to modulate the velocity network, but also to define a condition-dependent source distribution.

Despite this conceptual advantage, most existing flow matching methods still adopt a standard Gaussian source, which carries no information about the target distribution. This choice is largely inherited from diffusion models and motivated by simplicity rather than by the structure of the conditional generation problem. Recent work has begun to revisit this default from two complementary directions. One line improves the coupling between a Gaussian source and the data, often through optimal transport~\cite{tong2023otcfm, ot-multisample, cheng2025c2ot}; another modifies the source itself by constructing or learning a condition-dependent source~\cite{issachar2025designing, chen2025carflow, liu2025crossflow, he2025flowtok}.

Together, these works show that the source distribution can encode useful structure. However, for modern conditional flow matching, especially text-to-image generation, a systematic understanding remains incomplete. OT-based couplings clarify the role of transport ambiguity, but their benefits are difficult to realize in high-dimensional conditional regimes and may not transfer when applied naively~\cite{cheng2025c2ot}. Existing condition-dependent source designs demonstrate that conditioning can be injected into the source,
 but they offer limited guidance on \emph{why} a learned source reduces the difficulty of the flow-matching objective, \emph{how} to make source learning consistently beneficial in modern T2I regimes with high-dimensional latents~\cite{esser2024sd3, flux2024, zheng2025rae}, large-scale datasets~\cite{schuhmann2022laion, chen2025blip3o, russakovsky2015imagenet}, and strong conditioning~\cite{chen2025maetok, qin2025lumina, chen2023pixart}, or \emph{when} its benefits should be most pronounced.

In this work, we study condition-dependent source learning as a first-class design problem for conditional flow matching. We propose \emph{Condition-dependent Source Flow Matching} (\ours), a framework that jointly learns a condition-dependent source distribution and the velocity field under the flow-matching objective. Using modern text-to-image generation as a high-dimensional and strongly conditioned testbed, we address three central questions: \emph{why} source learning improves flow matching, \emph{how} it can be made stable and effective in practice, and \emph{when} its benefits are most pronounced. Our main contributions are:
\vspace{-0.5em}

\begin{itemize}[leftmargin=2em]
    \item \textbf{Why source learning helps.} We connect jointly learned condition-dependent sources to the intrinsic variance term of the flow-matching objective, which characterizes ambiguity in velocity supervision. Empirically, \ours{} provides cleaner supervision and improves optimization through lower gradient variance, straighter transport paths, and faster convergence.
    \vspace{-0.1em}
    \item \textbf{How to make source learning effective in modern T2I.} We identify key obstacles to stable source learning, including source collapse during joint optimization and weak supervision to the source generator in strongly conditioned T2I architectures. We show that variance-only regularization with directional source--target alignment enables stable end-to-end training without constraining source mobility.
    \vspace{-0.1em}
    \item \textbf{When source learning is most beneficial.} We show that the effectiveness of condition-dependent sources depends on the geometry of the target representation space. In particular, source learning yields larger gains in semantically structured target spaces, linking source design to recent advances in representations for generative modeling~\cite{zheng2025rae, yu2024repa, chen2025maetok, skorokhodov2025diffusability}.
    \vspace{-0.1em}
    \item \textbf{Empirical validation at modern T2I scale.} Through extensive experiments and ablations across multiple text-to-image benchmarks, \ours{} consistently improves over Gaussian-source baselines, achieving up to $\mathbf{3.01\times}$ faster FID convergence and $\mathbf{2.48\times}$ faster CLIP Score convergence, and outperforms representative condition-dependent source and coupling methods.

\end{itemize}

\section{Related work}
We discuss the most relevant related work here and provide additional details in Appx.~\ref{sec:appx_relwork}.

Recent works have revisited the standard Gaussian source in flow matching from two directions. 
The first modifies the source--target coupling while keeping the Gaussian source fixed. 
OT-based methods reduce path ambiguity through improved pairings~\cite{tong2023otcfm, ot-multisample}, and C$^2$OT extends this idea to conditional generation through condition-aware coupling~\cite{cheng2025c2ot}. 
However, these benefits are difficult to realize in high-dimensional conditional settings, where reliable minibatch OT estimation becomes challenging.

The second direction modifies the source distribution itself. 
CPD~\cite{issachar2025designing} constructs a condition-dependent prior through a separate regression objective decoupled from the FM loss, CAR-Flow~\cite{chen2025carflow} learns condition-dependent reparameterizations to align source and target distributions in simpler class-conditional settings, and CrossFlow and FlowTok~\cite{liu2025crossflow, he2025flowtok} demonstrate feasible text-dependent sources at T2I scale, but do not analyze source learning through FM training dynamics. Together, these works show that the source can encode conditioning information, but leave open a systematic understanding of \emph{why} source learning helps under the FM objective, \emph{how} to make source learning consistently beneficial beyond limited-scale T2I settings, and \emph{when} its benefits should be substantial.
\vspace{-0.5em}
\section{Preliminaries}
\label{sec:preliminaries}
\vspace{-0.5em}

\subsection{Flow Matching}
Flow Matching (FM) defines a generative process by solving an ordinary differential equation (ODE):
\begin{equation}
\frac{d}{dt} X_t = v_\theta(X_t, t), \quad t \in [0, 1],
\label{eq:ode}
\end{equation}
where $v_\theta(\cdot, t)$ is a neural vector field parameterized by $\theta$. The goal of FM is to find $\theta$ such that the push-forward of a source distribution $p_0$ through the ODE matches a target distribution $p_1$.

Typically, we consider a probability path $p_t$ and an associated marginal velocity field $u_t$ that generates $p_t$. For a given coupling $(X_0, X_1) \sim \pi$ with $X_0 \sim p_0$ and $X_1 \sim p_1$, we define the velocity target as:
\begin{equation}
\Delta := X_1 - X_0.
\label{eq:delta}
\end{equation}
Assuming a linear interpolation path $X_t = (1-t)X_0 + tX_1$, the marginal velocity field at a point $x$ and time $t$ is expressed as
\begin{equation}
u_t(x) = \mathbb{E}[\Delta \mid X_t = x].
\label{eq:marginal_u}
\end{equation}

The FM objective minimizes the mean squared error between the neural vector field and the velocity target:
\begin{equation}
\mathcal{L}_{\mathrm{FM}}(\theta) = \mathbb{E}_{t, \pi(X_0, X_1)} \big[ || v_\theta(X_t, t) - \Delta ||^2 \big],
\label{eq:fm_loss}
\end{equation}
where $t \sim p(t)$ is a probability distribution over $[0, 1]$. While each conditional path $t \mapsto X_t$ is linear, the learned marginal vector field generally induces curved trajectories because it aggregates multiple, potentially conflicting velocities $\Delta$ at the same spacetime point $(x,t)$.

For conditional flow matching, the target $X_1$ is paired with condition $C$, and the velocity field additionally takes $C$ as input. With the standard Gaussian source, one typically uses the condition-agnostic coupling $\pi(X_0,X_1,C)=p_0(X_0)p_1(X_1,C)$:
\begin{equation}
\mathcal{L}_{\mathrm{FM}}(\theta) = \mathbb{E}_{t, \pi(X_0, X_1, C)} \big[ || v_\theta(X_t, t, C) - \Delta ||^2 \big].
\label{eq:fm_loss_cond}
\end{equation}

\subsection{Bias-Variance decomposition of Flow Matching loss}
To analyze learning dynamics, the FM objective in Eq.~\eqref{eq:fm_loss} decomposes into two components (see Appx.~\ref{sec:app_fm_decomposition}):
\begin{align}
\mathcal{L}_{\mathrm{FM}}(\theta)
&= \underbrace{
\mathbb{E}_{t, \pi(X_0,X_1)} \Big[
\lVert v_\theta(X_t, t) - u_t(X_t) \rVert^2
\Big]
}_{\text{Approximation Error}}
+ \underbrace{
\mathbb{E}_{t, \pi(X_0,X_1)} \Big[
\mathrm{Var}(\Delta \mid X_t)
\Big]
}_{\text{Intrinsic Variance}} .
\label{eq:decomposition}
\end{align}

\begin{figure*}[t]
    \centering
    \includegraphics[width=1.0\linewidth]{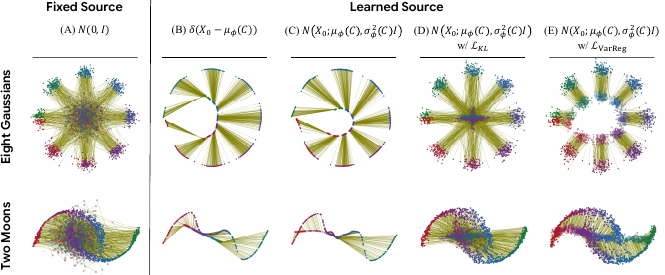}
    \caption{\textbf{Analysis of CSFM designs.} 
    We investigate the effect of the source designs using two two-dimensional synthetic datasets with continuous conditions: Eight Gaussians with polar angle condition and Two Moons with $x$-coordinate condition. We visualize the transport trajectories, where `$\boldsymbol{\times}$' denotes source points $X_0$ and `$\bullet$' denotes points $X_1^{\text{sampled}}$ generated by the flow model. Colors indicate the conditioning variable. \textbf{(A) Fixed Standard Gaussian: } Independent coupling results in entangled paths and high intrinsic variance. \textbf{(B) Deterministic Mapping: } The flow model with a deterministically mapped source fails to reconstruct the original target distribution. \textbf{(C) Conditional Gaussian: } Although the source is modeled as a condition-dependent Gaussian, its variance collapses during training, resulting in insufficient support and an inability to recover the target distribution. \textbf{(D) Conditional Gaussian with Standard KL regularization:} While preventing collapse to a deterministic mapping, the constraint on $\mu_\phi(C)$ limits the mobility of the source, yielding entangled trajectories. \textbf{(E) Conditional Gaussian with VarReg:} Variance-only regularization prevents collapse while allowing the conditional mode $\mu_\phi(C)$ to move, resulting in a target-aligned source distribution and disentangled trajectories.
    }
    \label{fig:toy2d}
    \vspace{-1.5em}
\end{figure*}

The Approximation Error measures how well the model recovers the marginal velocity field $u_t$, while the Intrinsic Variance is an irreducible term determined by the coupling $\pi$. For linear interpolants, this variance characterizes path ambiguity: it vanishes when $\Delta$ is uniquely determined by $X_t$, or equivalently when interpolants do not intersect~\cite{ma2025learningstraightflowsvariational}. Thus, high intrinsic variance indicates that multiple trajectories crossing the same spacetime point $(x,t)$ provide conflicting velocity supervision. 

Prior works show that reducing intrinsic variance improves coupling quality, leading to faster convergence and more stable flow learning~\cite{tong2023otcfm, ot-multisample}. OT-based methods affect this term by explicitly changing sample-level source--target pairings. However, explicit OT matching becomes unreliable in high-dimensional conditional settings, limiting their practical effectiveness~\cite{cheng2025c2ot}.

\vspace{-0.25em}
 \vspace{-0.5em}
\section{Method}
\label{sec:method}
 \vspace{-0.5em}
In this section, we present \emph{Condition-dependent Source Flow Matching} (\ours), which jointly learns a source distribution and the velocity field under the flow-matching objective (Fig.~\ref{fig:main_1column}). We formulate source learning as adaptive source--target coupling, then introduce two components for stable training: variance-only regularization to prevent source collapse while preserving source mobility, and directional source--target alignment to improve source learning under complex conditioning.

 \vspace{-0.5em}
\subsection{Learning condition-dependent sources under flow matching objective}

In conditional generation scenarios such as text-to-image generation, the conditioning variable $C$ is naturally paired with the data random variable $X_1$. 
We leverage this relationship by replacing the condition-agnostic Gaussian source in Eq.~\eqref{eq:fm_loss_cond} with a learnable condition-dependent source distribution $p_\phi(X_0 \mid C)$.
This defines a learnable condition-mediated coupling
$\pi_\phi(X_0, X_1, C) = p_\phi(X_0 \mid C)p_1(X_1, C)$,
where the source marginal adapts to the condition.

Specifically, we introduce a source generator $g_\phi(\cdot)$ that maps the condition $C$ to the parameters of $p_\phi(X_0 \mid C)$. We sample $X_0 \sim p_\phi(X_0 \mid C)$ and jointly train the flow model $v_\theta(\cdot)$ and the source generator $g_\phi(\cdot)$ under the conditional FM loss:
\begin{equation}
\begin{aligned}
\mathcal{L}_{\mathrm{FM}}(\theta, \phi)
&=
\mathbb{E}_{t, \pi_\phi(X_0, X_1, C)}
\left[
\left\lVert
v_\theta(X_t, t, C) - \Delta
\right\rVert^2
\right], \\
\text{where } X_0 &\sim p_\phi(X_0\mid C),\quad
\Delta = X_1 - X_0,\quad
X_t = (1 - t)X_0 + tX_1 .
\end{aligned}
\label{eq:fm-ours}
\end{equation}



By making the source--target coupling learnable through $p_\phi(X_0\mid C)$, \ours{} exposes the induced coupling to the FM objective rather than treating it as fixed. Since intrinsic variance is coupling-dependent (Eq.~\ref{eq:decomposition}), source learning can reduce ambiguity in the velocity target, as verified in toy settings in Appx.~\ref{subsec:appx_toy_intrinsic_variance}. This adaptive coupling yields fewer path intersections and cleaner velocity supervision, manifesting as lower gradient variance, faster convergence, and straighter flows. We visualize this effect in Fig.~\ref{fig:toy2d} and provide quantitative evidence in Sec.~\ref{subsec:advantages_of_csfm}.

\vspace{-0.25em}


\subsection{Preventing source collapse while preserving source mobility}
\label{subsec:design}
While a learnable conditional source can reduce velocity ambiguity, careful design is required.
\vspace{-0.25em}



\paragraph{Conditional Gaussian for sufficient support.} 
A straightforward source design is a deterministic mapping,
$p_\phi(X_0 \mid C) = \delta(X_0 - \mu_\phi(C))$.
However, this choice severely restricts the support\footnote{The support of a random variable $X$ is defined as the smallest closed set $S \subseteq \mathbb{R}^d$ such that $\mathbb{P}(X \in S) = 1$.}
of the source $X_0$.
As established by \cite{lee2025there}, an overly concentrated source causes path entanglement and degrades flow matching performance.
We empirically demonstrate this failure mode for deterministic conditional sources in toy experiments (Fig.~\ref{fig:toy2d} (B)), where the flow model fails to recover the distribution of $X_1$. 
To address this, we instantiate $p_\phi(X_0 \mid C)$ as a conditional Gaussian:
\vspace{0.1em}
\begin{equation}
p_\phi(X_0 \mid C)
=
\mathcal{N}\!\left(\mu_\phi(C),\, \operatorname{diag}(\sigma_\phi^2(C))\right),
\label{eq:source_gaussian}
\vspace{0.1em}
\end{equation}
which provides full support when $\sigma^2_\phi(C)>0$ and allows reparameterizable sampling~\cite{kingma2022autoencodingvariationalbayes}.

\vspace{-0.25em}
\paragraph{Variance regularization for collapse prevention.}
A learnable Gaussian source can adapt both its conditional location and stochastic support, but
joint training with the flow model often drives the source variance $\sigma_\phi^2(C)$ toward zero (Fig.~\ref{fig:toy2d}~(C)), since shrinking the source variance directly reduces the intrinsic variance in Eq.~\eqref{eq:decomposition}.
To counteract this effect, a common strategy for Gaussian parameterization is to regularize the distribution toward a standard normal prior~\cite{liu2025crossflow, he2025flowtok}. However, such regularization constrains both the variance and mean, forcing $\mu_\phi(C)$ toward the origin, which we find unnecessarily restricts the flexibility of the source distribution and degrades performance. As illustrated in Fig.~\ref{fig:toy2d}~(D), this constraint prevents the source from relocating toward target modes, resulting in entangled transport paths. To avoid this limitation, we adopt a variance-only regularization that penalizes deviations of $\sigma_\phi^2(C)$ from unit variance while leaving the mean $\mu_\phi(C)$ unconstrained: 
\begin{equation}
\small
\begin{aligned}
    \mathcal{L}_{\mathrm{VarReg}}(\phi)
    = \mathbb{E}_{C} \Big[
    D_{\mathrm{KL}}\big(
    \mathcal{N}(\mu_\phi(C), \operatorname{diag}(\sigma_\phi^2(C)))
    \Vert
    \mathcal{N}(\mu_\phi(C), \mathbf{I})
    \big)
    \Big]
\end{aligned}
\label{eq:var_reg}
\end{equation}

By leaving $\mu_\phi(C)$ unconstrained while anchoring the variance, the source distribution can relocate toward condition-compatible target regions without collapsing its support (Fig.~\ref{fig:toy2d} (E)). This avoids the mean-constraining effect of standard KL regularization and yields simpler, less entangled transport geometry with cleaner supervision signals.

\vspace{-0.25em}
\subsection{Strengthening source supervision under complex conditioning}
\label{subsec:learning}
Unlike simplified toy settings, practical conditional generation tasks such as text-to-image synthesis involve highly complex and multimodal relationships between the condition $C$ and the target data $X_1$. Modern text-to-image architectures~\cite{esser2024sd3, flux2024, qin2025lumina} are specifically designed to model such complexity by tightly integrating the conditioning signal into the flow model $v_\theta(\cdot)$, allowing the conditional information to directly modulate the vector field itself. However, this tight integration also makes the optimization of a suitable source distribution more challenging. Since the flow model can account for much of the conditional information through such modulation, minimizing the flow matching objective imposes weaker supervision signals for the source distribution, making it harder to learn an informative source in practice (see Appx.~\ref{sec:appx_balance}  for a detailed analysis).



To fully leverage modern conditional architectures while mitigating this optimization challenge, we introduce an explicit source--target alignment objective. Motivated by recent findings that directional information is critical in high-dimensional flow matching~\cite{lee2025there}, we adopt a negative cosine similarity loss to encourage directional alignment between learned source and target samples, providing source-specific guidance without directly matching target magnitudes or shrinking stochastic support:
\vspace{-0.1em}
\begin{equation}
\mathcal{L}_{\mathrm{align}}(\phi)
= \mathbb{E}_{C,X_1C}\!\left[\, 1 - \frac{ X_0 \cdot X_1 }{ \lVert X_0 \rVert \, \lVert X_1 \rVert } \right].
\end{equation}
Combined with the flow matching loss $\mathcal{L_\mathrm{FM}}$ and the variance regularization $\mathcal{L}_{\mathrm{VarReg}}$, the final training objective is:
\begin{equation}
\mathcal{L}_{\mathrm{total}} = \mathcal{L}_{\mathrm{FM}} + \lambda_{\mathrm{VarReg}} \mathcal{L}_{\mathrm{VarReg}} + \lambda_{\mathrm{align}} \mathcal{L}_{\mathrm{align}},
\end{equation}
where $\lambda_{\mathrm{VarReg}}$ and $\lambda_{\mathrm{align}}$ are hyperparameters that balance the distribution support and alignment quality of the learned source distribution, respectively. 
\vspace{-0.5em}

\newcommand{\rot}[1]{\rotatebox{0}{#1}}

\begin{table*}[t]
\centering
\caption{\textbf{Component-wise analysis on ImageNet 256$\times$256.} We analyze the effect of individual components on the captioned ImageNet-1K dataset. All models are trained for 100K iterations with a batch size of 1024 and evaluated using a 50-step Euler ODE sampler without guidance. Baseline models with a fixed Gaussian source are highlighted in \graybox{gray}, \textbf{bold} entries denote the default setting for subsequent experiments, and \underline{underlined} values indicate the best results. In this experiment, RAE (DINOv2) is used as the image autoencoder. ($^\dagger$ denotes a baseline model with increased parameters to approximately match the parameter count introduced by the source generator.)}
\label{tab:abl}
\resizebox{0.93\textwidth}{!}{
\begin{tabular}{c|cccc|cccccccc}
\toprule
\begin{tabular}[b]{@{}c@{}}Source \\ Distribution \end{tabular} &
\begin{tabular}[b]{@{}c@{}}\textbf{Text}\\\textbf{Encoder}\end{tabular} & 
\begin{tabular}[b]{@{}c@{}}\textbf{Flow}\\\textbf{Backbone}\end{tabular} & 
\begin{tabular}[b]{@{}c@{}}\textbf{Align}\\\textbf{Loss}\end{tabular} &  
\begin{tabular}[b]{@{}c@{}}\textbf{Reg.}\\\textbf{Loss}\end{tabular} &
\rot{FID~$\downarrow$} & \rot{CLIP~$\uparrow$} & \rot{IS$\uparrow$} & \rot{sFID$\downarrow$} & \rot{Prec.$\uparrow$} & \rot{Recall$\uparrow$} & \rot{FDD$\downarrow$} & \rot{VQAScore$\uparrow$}\\
\midrule
    \rowcolor{gray!10}  & & LightningDiT &  &  & 3.721 & 0.3283 & 169.2  & 6.175 & 0.7977 & 0.5718 & 87.65 & 0.2130  \\
    \rowcolor{gray!10} & & MMDiT &  &  & 3.412 & 0.3399 & 186.1 & 6.507 & 0.7906 & 0.5747 & 78.39 & 0.4710 \\
    \rowcolor{gray!10} & & UnifiedNextDiT &  &  & 3.036 & 0.3398 & 187.0  & 5.859 & 0.7917 & 0.5881 & 69.08 & 0.4819\\
    \rowcolor{gray!10} 
    \multirow{-4}{*}{\begin{tabular}[b]{@{}c@{}} $\mathcal{N}(0,I)$ \end{tabular}} &
    \multirow{-4}{*}{CLIP} & UnifiedNextDiT$^\dagger$ & 
    \multirow{-4}{*}{\blackx} & 
    \multirow{-4}{*}{\blackx} & 2.925 & 0.3396 & 189.1  & 5.730 & 0.7898 & 0.5974 & 65.29 & 0.4797\\
    
\midrule
    \multirow{12}{*}{\begin{tabular}[b]{@{}c@{}} $p_\phi(X_0\mid C)$  \end{tabular}} & 
    \multirow{4}{*}{CLIP} & \multirow{3}{*}{UnifiedNextDiT} & & \cellcolor{hlgreen}{\blackx} & NaN & NaN & NaN & NaN & NaN & NaN & NaN & NaN  \\
    & & & & \cellcolor{hlgreen}{KL} & 2.904 & \underline{0.3405} & 190.1 & 5.671 & 0.7869 & 0.5913 & 65.92 & \underline{0.4854} \\
    & & & \multirow{-3}{*}{\blackx} & \cellcolor{hlgreen}{\textbf{VarReg}}  & \underline{2.765} & 0.3404 & \underline{195.6} & \underline{5.630} & \underline{0.7921} & \underline{0.5958} & \underline{60.64} & 0.4747  \\

\cmidrule(lr){2-13}
    & \multirow{3}{*}{CLIP} & \multirow{3}{*}{UnifiedNextDiT} & \cellcolor{hlblue}{\blackx} &  & 2.765 & 0.3404 & 195.6 & 5.630 & 0.7921 & 0.5958   & 60.64 & 0.4747 \\
    & & & \cellcolor{hlblue}{MSE} &  & 2.942 & 0.3410 & 195.5 & 6.247 & 0.7896 & 0.5902 & 69.89 & 0.5006 \\
    & & & \cellcolor{hlblue}{\textbf{CosSim}} & \multirow{-3}{*}{VarReg} & \underline{2.453} & \underline{0.3420} & \underline{203.7} & \underline{5.491} & \underline{0.7947} & \underline{0.6029} & \underline{54.00} & \underline{0.5105} \\
\cmidrule(lr){2-13}
     & \multirow{3}{*}{CLIP} & \cellcolor{hlpurple}{LightningDiT} & \multirow{3}{*}{CosSim} & \multirow{3}{*}{VarReg} & 3.041 & 0.3363 & 191.9 & 5.972 & 0.7961 & 0.5791 & 68.64 & 0.3283 \\
     & & \cellcolor{hlpurple}{MMDiT} & & & 3.051 & 0.3415 & 199.0 & 6.339 &  \underline{0.7978} & 0.5745 & 69.64 & 0.4994\\
     & &  \cellcolor{hlpurple}{\textbf{UnifiedNextDiT}} & & &  \underline{2.453} &  \underline{0.3420} &  \underline{203.7} &  \underline{5.491} & 0.7947 &  \underline{0.6029} & \underline{54.00} & \underline{0.5105} \\
\cmidrule(lr){2-13}
    & \cellcolor{hlred}{\textbf{CLIP}} & \multirow{2}{*}{UnifiedNextDiT} &  &  &  \underline{2.453} &  \underline{0.3420} &  \underline{203.7} &  5.491 & 0.7947 &  \underline{0.6029} & \underline{54.00} & 0.5105  \\
    & \cellcolor{hlred}{Qwen3} &  & \multirow{-2}{*}{CosSim} & \multirow{-2}{*}{VarReg}& 2.519 & 0.3409 & 200.5 &  \underline{5.465} &   \underline{0.7953} & 0.6009 & 54.84 & \underline{0.5215} \\

\bottomrule
\end{tabular}}\vspace{-10pt}
\end{table*}

\vspace{-0.25em}
\section{Experiments}
\label{sec:experiments}
\vspace{-0.25em}
In this section, we first validate the design components that make source learning stable and effective in practice (\emph{how}), then demonstrate the underlying mechanism through training dynamics and flow geometry (\emph{why}), and finally identify the representational conditions under which source learning is most effective (\emph{when}). We conclude by scaling \ours\ to a 1.3B model on a modern T2I benchmark.

\vspace{-0.5em}
\paragraph{Implementation details.}
We use RAE~(DINOv2)~\cite{zheng2025rae} as the default target image representation and adopt a DDT head~\cite{wang2025ddt} to provide sufficient model capacity for the high-dimensional feature space.
The motivation for the target representation choice, along with a deeper analysis, is discussed in Sec.~\ref{subsec:rep_analysis}.
To facilitate fair evaluation at an appropriate scale for practical text-to-image generation tasks, we construct a benchmark dataset based on ImageNet-1K~\cite{russakovsky2015imagenet}. To study T2I generation in a standardized setting, we employ Qwen3-VL~\cite{Qwen3-VL} to generate descriptive captions for images (see Fig.~\ref{fig:appx_dataset}), rather than simple class-level captions~\cite{degeorge2025far}.
Model performance is primarily evaluated using FID~\cite{heusel2017fid} and CLIP Score~\cite{hessel2021clipscore}, together with Inception Score (IS)~\cite{salimans2016inceptionscore}, sFID, Precision--Recall, FDD~\cite{stein2023fdd}, and VQAScore~\cite{lin2024vqascore} metrics on the validation set.
For ImageNet-1K evaluation, images are generated from validation captions and evaluated on the full 50K-image validation set.
Unless otherwise specified, all evaluations are conducted without any guidance; guidance-related results and details are provided in Appx.~\ref{sec:appx_guidance}.
Additional training and evaluation details are provided in Appx.~\ref{sec:appx_train}.


\vspace{-0.5em}
\subsection{Validating design components}
\label{subsec:main_result}
\vspace{-0.5em}
\paragraph{\colorbox{hlgreen}{Choice of source regularization matters.}} We examine the role of the regularization loss $\mathcal{L}_{\mathrm{VarReg}}$ in stabilizing source learning. Using CLIP~\cite{radford2021learning} as the text encoder, we evaluate how variance regularization affects model behavior. Without any regularization, the learned source distribution collapses, driving the training objective near zero and preventing meaningful generation. Standard KL regularization avoids this collapse and improves upon a fixed Gaussian source, but the gains remain limited. As discussed in Sec.~\ref{subsec:design}, this is due to the KL regularization constraining the source mean $\mu_\phi(C)$ toward the origin, which restricts source mobility. In contrast, our variance regularization preserves variance control while allowing the mean to adapt freely, enabling effective alignment with the target distribution. This design already yields broad improvements across evaluation metrics, highlighting the importance of unconstrained mean adaptation for learning expressive source distributions in large-scale text-to-image generation.

\begin{figure}[t]
\centering

\begin{minipage}{0.48\textwidth}
    \centering
    \includegraphics[width=\linewidth]{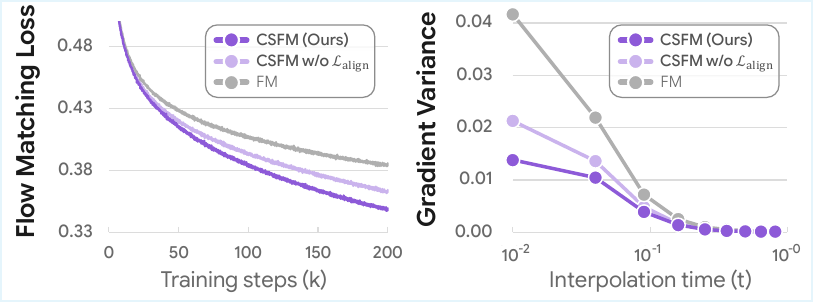}
    \caption{\textbf{Flow matching loss and gradient variance ($\mathrm{Var}(\nabla_\theta \mathcal{L}_{\text{FM}})$).} 
    We compare the training dynamics of standard FM, CSFM without alignment loss, and CSFM. 
    CSFM achieves faster loss convergence and lower gradient variance, particularly at early interpolation times near the source. Details of the measurement are provided in Appx.~\ref{subsec:appx_gradvar}.}
    \label{fig:fmloss_gradvar}
\end{minipage}
\hspace{0.02\textwidth}
\begin{minipage}{0.48\textwidth}
    \centering
    \includegraphics[width=\linewidth]{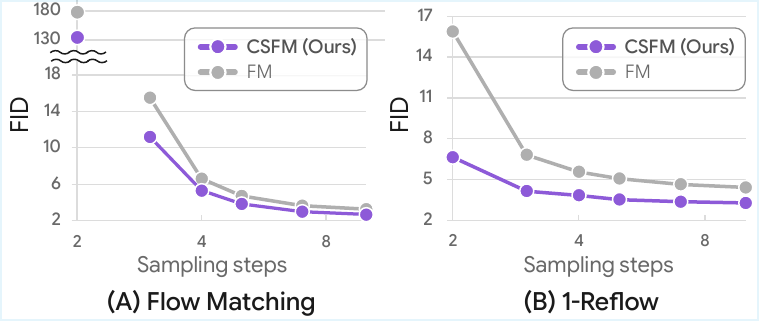}
    \caption{\textbf{Few-step generation and flow straightness.} 
    We compare FID across different sampling steps for (A) Flow Matching and (B) 1-Reflow. CSFM degrades more gracefully as the number of steps decreases, indicating reduced path intersections and a straighter transport field compared to the FM baseline.}
    \label{fig:fewstep_fid}
\end{minipage}

\vspace{-1em}
\end{figure}

\paragraph{\colorbox{hlblue}{Alignment is effective in complex settings.}} We next examine the role of the alignment loss $\mathcal{L}_{\mathrm{align}}$ in improving optimization when learning condition-dependent sources under complex conditioning. As discussed in Sec.~\ref{subsec:learning}, strong conditioning in the flow model $v_\theta(\cdot)$ can substantially weaken the learning signal available to the source, making optimization unstable and often yielding a poorly trained source distribution (see Appx.~\ref{sec:appx_balance}).
In this setting, directional alignment substantially improves performance and mitigates these optimization difficulties. 
Importantly, CSFM without the alignment loss exhibits higher FM loss and increased gradient variance in Fig.~\ref{fig:fmloss_gradvar}, indicating that alignment improves flow matching optimization itself rather than acting as an external regularizer. A more detailed discussion of these statistics is provided in Sec.~\ref{subsec:advantages_of_csfm}. To further highlight the importance of directional alignment, we compare this approach with an MSE objective and find that directly minimizing the $\ell_2$ distance between $X_0$ and $X_1$ overly restricts the source distribution, limiting the flexibility needed for effective source learning under the FM objective.
\vspace{-0.5em}
\paragraph{\colorbox{hlpurple}{Robustness to conditioning architecture.}} 
We evaluate whether CSFM is consistently effective across different conditioning architectures.
We consider three representative paradigms in modern text-to-image generation: LightningDiT~\cite{yao2025lightning}, which injects conditioning primarily via adaptive layer normalization (AdaLN)~\cite{peebles2023dit}; MMDiT~\cite{esser2024sd3}, which adopts a dual-stream design that processes text and image tokens in separate but interacting branches; and UnifiedNextDiT~\cite{qin2025lumina}, which employs a unified sequence representation with Multimodal RoPE (mRoPE)~\cite{bai2025qwen25} to capture cross-modal structure without explicit modulation layers. 
As shown in Tab.~\ref{tab:abl}, our method consistently improves performance over the \colorbox{gray!10}{baseline} across all three architectures. These results indicate that \ours\ yields robust gains in text-to-image generation, independent of the specific conditioning mechanism used by the backbone.
\vspace{-0.5em}
\paragraph{\colorbox{hlred}{Robustness to text encoders.}}
We further assess the robustness of our method by replacing the CLIP text encoder with a large language model (LLM). Specifically, we use Qwen3-0.6B~\cite{yang2025qwen3} as the text encoder and observe that our method maintains comparable performance to the CLIP-based setting. This result indicates that our framework generalizes across different text encoder architectures and is not tied to a specific encoder design.

Unless otherwise specified, all subsequent experiments adopt the default configuration highlighted in \textbf{bold} in Tab.~\ref{tab:abl}.

\vspace{-0.5em}
\subsection{Why CSFM improves Flow Matching}
\label{subsec:advantages_of_csfm}
\paragraph{\ours\ improves training dynamics of flow matching. }  
By jointly learning a condition-dependent source under the FM loss, \ours{} can reduce the intrinsic variance term in Eq.~\eqref{eq:decomposition}, which we directly verify in toy settings in Appx.~\ref{subsec:appx_toy_intrinsic_variance}. This leads to markedly improved optimization dynamics for flow matching. As shown in Fig.~\ref{fig:fmloss_gradvar}, \ours\ achieves a faster decrease in the FM loss, indicating accelerated convergence~\cite{benton2023nearly, benton2023error}. 
Looking more closely at the optimization behavior, effective source--target coupling is known to reduce the gradient variance of the FM loss~\cite{ot-multisample}. We empirically examine this behavior by measuring the gradient variance at 100K training step in Fig.~\ref{fig:fmloss_gradvar}.
\ours\ consistently attains lower gradient variance than the baseline, with pronounced gains at small interpolation times (i.e., near the source). This further supports that the condition-dependent source provides cleaner velocity supervision and reduces the corresponding ambiguity.
\vspace{-0.25em}

\begin{figure}[t]
\centering

\begin{minipage}{0.48\textwidth}
  \centering
  \includegraphics[width=0.98\linewidth]{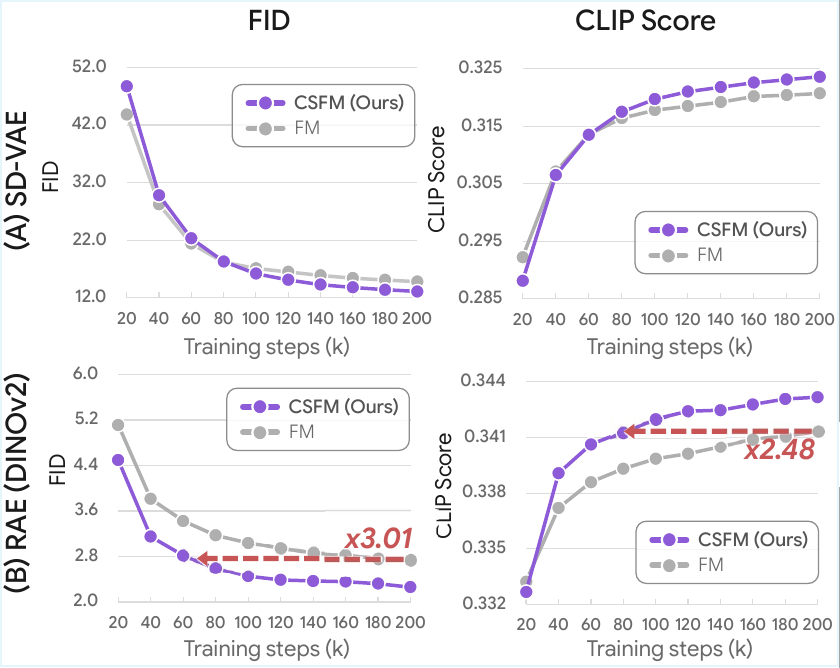}
  \caption{\textbf{Training efficiency under different target representations.} 
  We compare FID and CLIP Score trajectories between CSFM and FM, using (A) SD-VAE and (B) RAE (DINOv2) target representations on the ImageNet-1K validation set. While CSFM yields consistent gains under both representations, it substantially accelerates convergence and achieves larger improvements in the structured RAE space.}
  \label{fig:training_efficiency}
\end{minipage}
\hspace{0.02\textwidth}
\begin{minipage}{0.48\textwidth}
  \centering
  \includegraphics[width=0.98\linewidth]{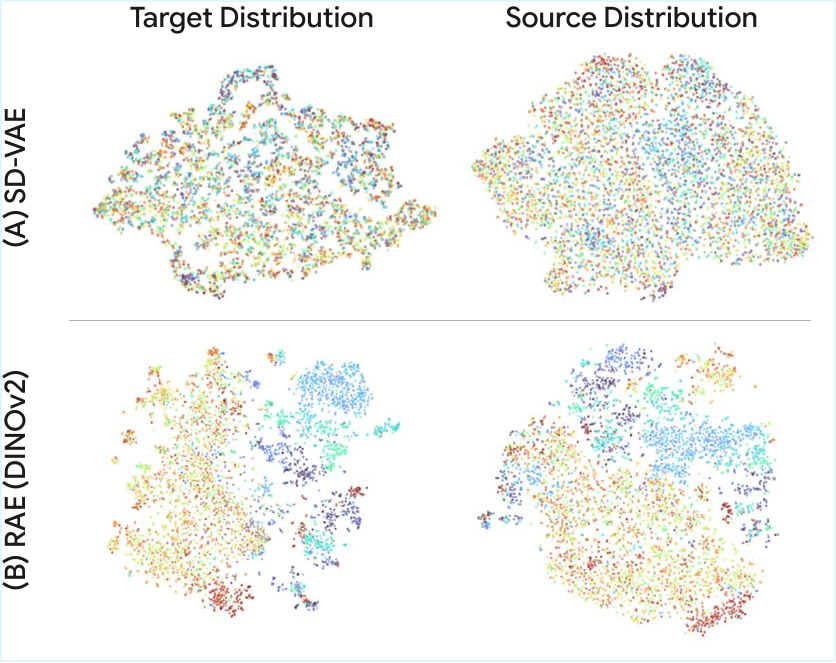}
  \caption{\textbf{t-SNE visualization of target and learned source distributions.} 
  We visualize t-SNE embeddings of target (left) and corresponding learned source (right) distributions, colored by class labels, for (A) SD-VAE and (B) RAE (DINOv2) representations. The entangled SD-VAE space leads to an equally entangled source, whereas the structured RAE space enables a semantically organized source distribution.}
  \label{fig:source_comparison}
\end{minipage}

\vspace{-1.5em}
\end{figure}

\paragraph{\ours\ improves flow matching performance. }
We show that the improved training dynamics translate into performance gains. 
As shown in Fig.~\ref{fig:training_efficiency}, \ours\ achieves consistent improvement, yielding a $3.01\times$ speedup in FID convergence and a $2.48\times$ speedup in CLIP Score in the RAE~\cite{zheng2025rae} latent space. We take a deeper look at the performance discrepancy between Stable Diffusion VAE (SD-VAE) and RAE in Sec.~\ref{subsec:rep_analysis}. 
Qualitatively, \ours\ tends to better reflect complex text conditioning involving multiple objects and relationships, while preserving high visual fidelity, as shown in Fig.~\ref{fig:appx_imagenet_qual_complexcaption}.

\vspace{-0.5em}

\vspace{-0.25em}
\paragraph{\ours\ straightens transport paths. }
Reducing the intrinsic variance in Eq.~\eqref{eq:decomposition} is known to minimize path intersections and induces straighter flow fields~\cite{tong2023otcfm,ma2025learningstraightflowsvariational}. We therefore evaluate flow straightness via few-step generation. As shown in Fig.~\ref{fig:fewstep_fid}~(A), \ours\ degrades more gracefully than standard FM as the number of sampling steps decreases: when reducing from 50 to 3 steps, FID degrades by $8.75$ for \ours\, compared to $12.47$ for the baseline.
To further investigate the potential straightness of the learned source distribution, we conduct 1-Reflow~\cite{liu2022rf} experiments. Reflow rectifies the flow fields and improves few-step generation, by fine-tuning the flow model on sampled pairs $(X_0, X_1^{\mathrm{sampled}})$. We fine-tune the flow models for 20K step from 100K-step checkpoint, while freezing the source generator for \ours.
As shown in Fig.~\ref{fig:fewstep_fid} (B), \ours\ exhibits substantially straighter flow fields: reducing from 50 to 2 sampling steps, FID degrades by only $3.51$, whereas standard FM suffers a larger degradation of $11.75$.
These results provide empirical evidence that the learned source distribution reduces path intersections and induces straighter flow fields.



\vspace{-0.5em}
\subsection{When source learning helps: Target representation matters}
\vspace{-0.5em}
\label{subsec:rep_analysis}

In Fig.~\ref{fig:training_efficiency}, \ours\ improves performance across image representation spaces, but with much larger gains in RAE (DINOv2)~\cite{zheng2025rae} latents than in SD-VAE~\cite{dhariwal2021diffusion}. This indicates that the structure of the target representation is critical for effective source learning. We further verify that this trend generalizes to another source-learning method in Appx.~\ref{sec:appx_rep_other_method}.


At a high level, learning a condition-dependent source is most beneficial when the conditioning signal $C$ induces a well-separated and discriminative structure in the target space, as illustrated in Fig.~\ref{fig:toy2d} and Fig.~\ref{fig:appx_toy2d_illcond}. In such cases, samples sharing the same condition form relatively compact clusters, making the source mean $\mu_\phi(C)$ well-defined and easier to align with the target distribution. This reduces path intersections, yielding more coherent flow-matching supervision.

However, this advantage diminishes when a single condition corresponds to a highly multimodal target distribution. As shown in Appx.~\ref{subsec:appx_ill}, when samples sharing the same $C$ are spread across distant modes, the source mean becomes ambiguous. This ambiguity leads to frequent path intersections, conflicting supervision, and persistently high intrinsic variance. In such cases, the learned source behaves more like a fixed Gaussian prior and provides limited additional benefit.

This property makes \ours\ most effective with representation autoencoders operating in structured latent spaces, such as DINOv2~\cite{oquab2023dinov2} or SigLIP2~\cite{tschannen2025siglip}. In these spaces, samples sharing the same condition tend to be more concentrated~\cite{huh2024platonic, wybitul2026representations, bolya2025perception}, reducing multimodality with respect to $C$. This concentration simplifies source learning and improves the effectiveness of \ours.


We further support this analysis with t-SNE visualizations in Fig.~\ref{fig:source_comparison}, where points are color-coded by their corresponding classes. In the SD-VAE representation, the target distribution exhibits strong entanglement and weak structure with respect to the conditioning signal, and this lack of structure is reflected in a similarly non-discriminative learned source (Fig.~\ref{fig:source_comparison} (A)). In contrast, the RAE (DINOv2) representation induces a more organized target geometry, which in turn allows the source distribution to become more discriminative and results in larger performance gains (Fig.~\ref{fig:source_comparison} (B)).
\vspace{-0.5em}

\begin{table*}[t]
\centering

\begin{minipage}[t]{0.49\textwidth}
\centering
\captionof{table}{\textbf{Comparison between condition-dependent source and coupling methods on ImageNet 256$\times$256.} 
We compare methods that learn a text-dependent source distribution or perform text-aware coupling. All baselines are reproduced adopting RAE~(DINOv2) for target latent using identical model architectures for fair comparison and evaluated with 50-step Euler ODE sampler.}
\vspace{0.5em}
\resizebox{\linewidth}{!}{
\begin{tabular}{l|ccccc}
\toprule
\textbf{Method} & \textbf{FID~$\downarrow$} & \textbf{CLIP~$\uparrow$} & 
\textbf{IS~$\uparrow$} & \textbf{FDD~$\downarrow$} &
\textbf{VQAScore~$\uparrow$}\\
\midrule
Standard FM & 3.036 & 0.3398 & 187.0 & 69.08 & 0.4819 \\
\midrule
CrossFlow~\cite{liu2025crossflow} & 2.957 & 0.3301 & 182.8 & 66.21 & 0.2613 \\
C$^2$OT~\cite{cheng2025c2ot} & 3.365 & 0.3406 & 184.3 & 78.72 & 0.4890 \\
CPD~\cite{issachar2025designing} & 3.219 & 0.3327 & 190.2 & 80.48 & 0.4113 \\
CAR-Flow~\cite{chen2025carflow} & 2.929 & 0.3408 & 192.8 & 66.38 & 0.4872 \\
\rowcolor{blue!10}
\textbf{CSFM (Ours)} & \textbf{2.453} & \textbf{0.3420} & \textbf{203.7} & \textbf{54.00} & \textbf{0.5105} \\
\bottomrule
\end{tabular}}
\label{tab:learned_source_comparison}
\end{minipage}
\hfill
\begin{minipage}[t]{0.47\textwidth}
\centering
\captionof{table}{\textbf{Large-scale text-to-image evaluation.} Results on GenEval and DPG-Bench. We report standard FM and CSFM with UnifiedNextDiT (1.3B), together with results from prior work that evaluate the same model families at different parameter scales, to contextualize the magnitude of performance gains at this scale.}
\vspace{0.4em}
\resizebox{\linewidth}{!}{
\begin{tabular}{l|c|cc}
\toprule
\textbf{Model} & \textbf{\#Params} & \textbf{GenEval} & \textbf{DPG-Bench} \\
\midrule
\multirow{2}{*}{BLIP3o~\cite{chen2025blip3o}} & 4B & 0.81 & 79.36 \\
& 8B & 0.84 & 81.60 \\
\multirow{2}{*}{Sana~\cite{xie2024Sana}} & 0.6B & 0.64 & 83.60 \\
& 1.6B & 0.66 & 84.80 \\
\midrule
Standard FM & 1.3B & 0.77 & 78.31 \\
\rowcolor{blue!10}
\textbf{CSFM (Ours)} & 1.3B & \textbf{0.80} & \textbf{81.11} \\
\bottomrule
\end{tabular}}
\label{tab:scale_results}
\end{minipage}

\vspace{-1em}
\end{table*}
\vspace{-0.25em}
\subsection{Comparison with condition-dependent source and coupling methods}
\vspace{-0.25em}
We compare \ours\ with C$^2$OT~\cite{cheng2025c2ot}, CPD~\cite{issachar2025designing}, CrossFlow~\cite{liu2025crossflow}, and CAR-Flow~\cite{chen2025carflow} in Tab.~\ref{tab:learned_source_comparison}, using the same RAE (DINOv2) target representation and UnifiedNextDiT architecture for fair comparison. Although these methods modify either the coupling or the source distribution, they provide limited gains in this high-dimensional T2I setting and often fail to improve over the Gaussian-source baseline. In contrast, \ours\ achieves the best performance across fidelity and alignment metrics. This suggests that, at modern T2I scale, the benefit of source conditioning depends critically on how the source is trained and regularized under the FM objective. Details of this comparison are provided in Appx.~\ref{sec:appx_impl_detail_comparison}.
\vspace{-0.5em}

\vspace{-0.25em}
\subsection{Scaling CSFM to modern text-to-image scale}
\vspace{-0.25em}
To examine whether \ours\ remains effective at scale, we scale the default configuration in Tab.~\ref{tab:abl} to a 1.3B-parameter model and replace the text encoder with Qwen3-0.6B for longer text inputs. The model is pretrained on the BLIP3o pretraining dataset~\cite{chen2025blip3o}, which contains approximately 36M samples, and then fine-tuned on BLIP3o-60K. We use a SigLIP2-based RAE decoder~\cite{tong2025scalerae} at $224\times224$ resolution and evaluate on GenEval~\cite{ghosh2023geneval} and DPG-Bench~\cite{hu2024ella}. As shown in Tab.~\ref{tab:scale_results}, \ours\ consistently outperforms the Gaussian-source baseline across benchmarks, demonstrating that learnable source distributions remain effective for high-capacity text-to-image generation. Full benchmark results are provided in Appx.~\ref{sec:appx_full_benchmark}.


However, as standard text-to-image benchmarks become increasingly saturated at this scale, quantitative metrics provide only a limited view of model behavior. We therefore present qualitative comparisons in Fig.~\ref{fig:main_1column}, Fig.~\ref{fig:appx_qual1}, and Fig.~\ref{fig:appx_qual2}, illustrating perceptual differences induced by source design and providing complementary evidence for the benefits of our approach in large-scale settings.

\vspace{-0.5em}
\section{Conclusion}
\vspace{-0.5em}
In this work, we present \emph{Condition-dependent Source Flow Matching} (\ours), treating the source distribution as a learnable design choice for conditional flow matching. We show \emph{why} source learning helps through the lens of intrinsic-variance-related ambiguity, \emph{how} stable end-to-end training can be achieved with variance-only regularization and directional alignment, and \emph{when} the gains are most pronounced in semantically structured target spaces. Extensive experiments demonstrate faster convergence, straighter flows, and consistent improvements at modern text-to-image scale. 

{
    \small
    \bibliographystyle{plainnat}
    \bibliography{neurips_2026}
}
\clearpage


\appendix
\section*{\Large Appendix}

\section{Toy experiments}
\label{sec:appx_toy}
\subsection{Toy experiments setup}
\label{subsec:appx_toy_exp_setup}
We conduct toy experiments (Fig.~\ref{fig:toy2d}) on two standard two-dimensional synthetic datasets, Eight Gaussians with polar angle condition and Two Moons with $x$-coordinate condition, using the official C$^2$OT implementation~\cite{cheng2025c2ot}. For the conditional flow model, the condition, time and the coordinate $x_t$ are projected into a shared hidden dimension, summed and passed through a 64-dimensional, 10-layer MLP with GELU activations. For the source generator, we use a 3-layer MLP, also with GELU activations. We train the network for 20K optimization steps with a batch size of 256, using Adam~\cite{2015-kingma-adam} with a learning rate of $3\times10^{-4}$. For visualization in Fig.~\ref{fig:toy2d}, we generate samples using an Euler sampler with 16 integration steps.

\subsection{Intrinsic variance comparison}
\label{subsec:appx_toy_intrinsic_variance}

\begin{figure}[h]
    \centering
    \includegraphics[width=0.7\linewidth]{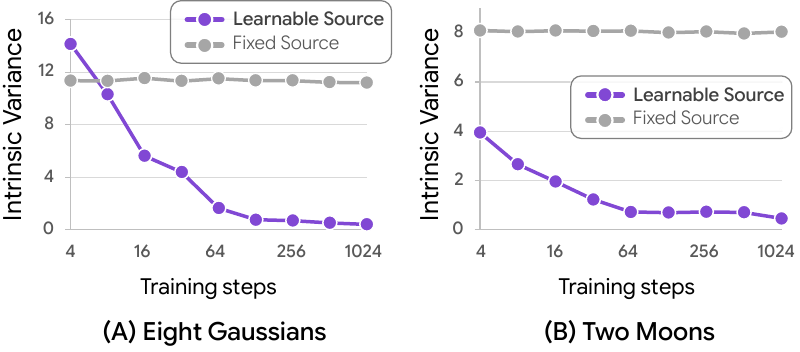}
    \caption{
       \textbf{Estimated intrinsic variance} for fixed Gaussian source (Fig.~\ref{fig:toy2d} (A)) and learnable source (Fig.~\ref{fig:toy2d} (E)). }
    \label{fig:appx_toy_intrinsic_variance}
\end{figure}

We estimate the intrinsic variance $\mathbb{E}[\mathrm{Var}(\Delta \mid X_t)]$ during training for both the fixed Gaussian source flow model and the learnable source flow model, as shown in Fig.~\ref{fig:appx_toy_intrinsic_variance}. Under a fixed Gaussian source, this term is determined by the prescribed coupling and remains essentially unchanged in expectation during training. In contrast, jointly training the source and flow model makes the coupling adaptive under the flow-matching objective, reducing the intrinsic variance over training.


\paragraph{Intrinsic variance estimation.}
We directly estimate the intrinsic variance term $\mathbb{E}[\mathrm{Var}(\Delta \mid X_t)]$, where $\Delta = X_1 - X_0$ and $X_t = (1-t)X_0 + tX_1$, in the toy experiments. At each evaluation checkpoint, we construct a pool of $N=100{,}000$ source--target pairs using the same coupling as in training, sample $t \sim \mathrm{Uniform}(0,1)$, and compute the corresponding $(X_t,\Delta)$. Since exact conditioning on the same $X_t$ is not possible in continuous space, we approximate the local conditional distribution using nearest neighbors in $X_t$-space. 
Specifically, for $M=10{,}000$ query points, we define $\mathcal{N}_K(i)$ as the set of the $K=64$ nearest neighbors of the $i$-th query point in the pool, measured in $X_t$-space and excluding the query point itself, and compute

\[
\widehat{\mathrm{IV}}
=
\frac{1}{M}
\sum_{i=1}^{M}
\frac{1}{K}
\sum_{j \in \mathcal{N}_K(i)}
\left\|
\Delta_j - \bar{\Delta}_i
\right\|_2^2,
\qquad
\bar{\Delta}_i
=
\frac{1}{K}
\sum_{j \in \mathcal{N}_K(i)}
\Delta_j .
\]
This estimates the average local variance of the velocity targets around each intermediate point $X_t$.

\subsection{Condition-dependent source with an unconditional flow model}
\label{subsec:appx_complex}

We conduct additional toy experiments using an \emph{unconditional} flow model to investigate how the learnable condition-dependent source distribution is influenced by conditioning injected into the flow model. Specifically, we compare a conditional source trained with an unconditional flow model $v_\theta(X_t,t)$ against one trained with a conditional flow model $v_\theta(X_t,t,C)$. Note that alignment loss in Sec.~\ref{subsec:design} is not applied for toy experiments.

\begin{figure}[h]
    \centering
    \includegraphics[width=0.5\linewidth]{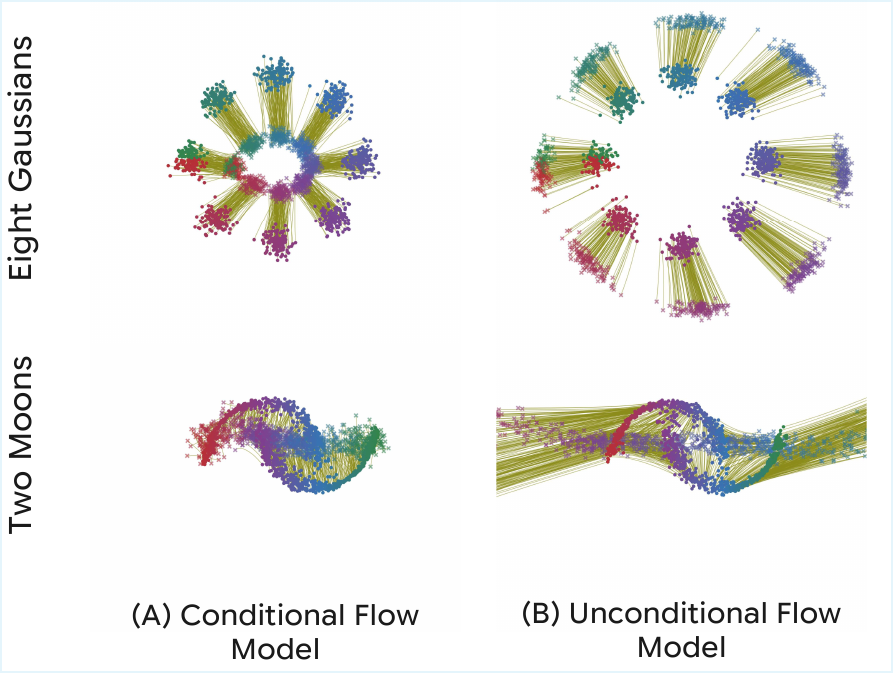}
    \caption{
       \textbf{Learned conditional source distributions with conditional vs.\ unconditional flow models.}}
    \label{fig:appendix_toy2d_uncondflow}
\end{figure}

As shown in Fig.~\ref{fig:appendix_toy2d_uncondflow} (B), when the flow model does not receive the condition $C$, the learned source distribution becomes noticeably more discriminative across conditions. This is mainly because in this setting, the source generator must encode condition-specific structure in order to reduce the flow matching loss, resulting in a more informative and well-separated conditional source. However, with unconditional flow model, the vector field is shared across conditions, making it harder for transport trajectories to branch at similar spatial locations.


This experiment should be viewed as a diagnostic for the role of the learned source. In the unconditional-flow variant, the condition can influence the velocity target only through the sampled source $X_0 \sim p_\phi(X_0\mid C)$ and the resulting intermediate state $X_t=(1-t)X_0+tX_1$, making the reduction of source-induced ambiguity directly visible in the learned source. In the main conditional model, $v_\theta(X_t,t,C)$ also receives $C$, so part of the conditional structure can be absorbed by the velocity network, weakening the learning signal to the source. However, with finite-capacity networks and continuous high-dimensional conditions, making $X_t$ itself condition-compatible can still ease the supervision seen by the flow model. This motivates the explicit alignment objective in Sec.~\ref{subsec:learning}, which encourages the source to retain useful condition-dependent structure during joint training. We further illustrate this optimization effect in a practical setting in Appx.~\ref{sec:appx_balance}.

\subsection{Ill-conditioned cases}
\label{subsec:appx_ill}
\begin{figure*}[h]
    \centering
    \includegraphics[width=0.5\linewidth]{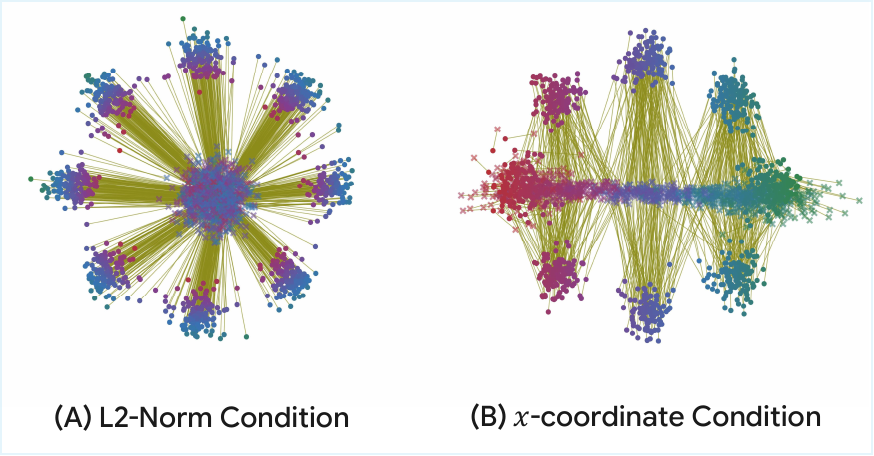}
    \caption{\textbf{Learned conditional source distributions under ill-conditioned settings.}}
    \label{fig:appx_toy2d_illcond}
\end{figure*}

We further investigate ill-conditioned scenarios in Fig.~\ref{fig:appx_toy2d_illcond}, where the conditioning signal does not sufficiently concentrate the target distribution. Specifically, we consider two extreme cases in which the condition is defined by the $\ell_2$ norm of the modes or by the $x$-coordinate in the Eight Gaussians setting. In these cases, a single condition value can correspond to multiple spatial locations, making the placement of a suitable source inherently ambiguous. Consequently, the learned source gravitates toward an averaged location—for instance, the midpoint between modes under the $x$-coordinate condition, or the center of the distribution in the extreme $\ell_2$-norm case—resulting in a more Gaussian-like source.

Despite still encouraging straighter transport paths compared to a fixed Gaussian prior, the learned source distribution in this setting becomes less discriminative across conditions, effectively reverting toward a Gaussian-like baseline (Fig.~\ref{fig:appx_toy2d_illcond} (A)). This observation aligns with the representation analysis in Sec.~\ref{subsec:rep_analysis}: learnable, condition-dependent source distributions are most effective when the target representation provides sufficient structure to reduce multimodality with respect to the condition.

\subsection{Variance explosion cases: stopping gradient on target velocity}
In the context of self-supervised learning~\cite{grill2020bootstraplatentnewapproach,caron2021emergingpropertiesselfsupervisedvision,xu2025nextembeddingpredictionmakesstrong}, stopping gradients is a commonly adopted strategy to prevent representational collapse. Motivated by this practice, one may consider stopping the gradient on the target velocity in Eq.~\ref{eq:fm-ours} when jointly learning the flow model and the source generator. Concretely, this corresponds to treating the target velocity $\Delta = \textrm{sg(}X_1 - g_\phi(C)\textrm{)}$ as a constant with respect to the source parameters, where $\textrm{sg}$ denotes the stop-gradient operation.

\begin{figure*}[h]
    \centering
    \includegraphics[width=0.5\linewidth]{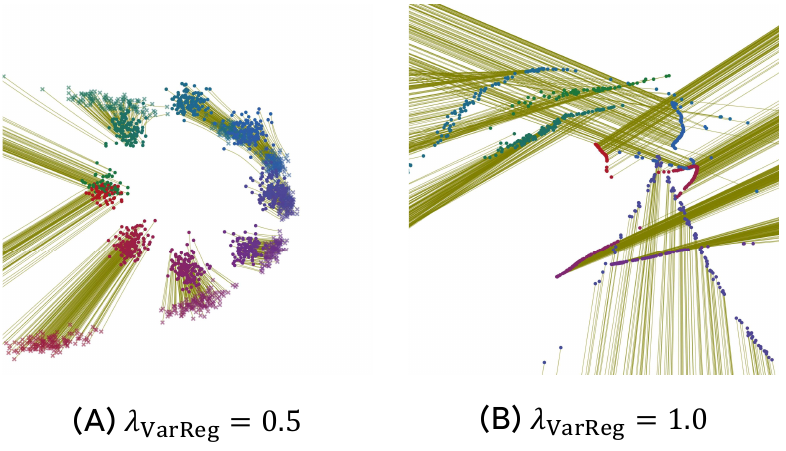}
    \caption{
       \textbf{Variance explosion case.  }The variance of the learned source distribution diverges under an unconditional flow model for (B) $\lambda_{\mathrm{VarReg}}=1.0$.
    }
    
    \label{fig:appendix_toy2d_stopgrad_explode}
\end{figure*}

However, as shown in Fig.~\ref{fig:appendix_toy2d_stopgrad_explode}, we find that stopping gradients can cause the explosion of the variance of the learnable source distribution in some cases. This behavior corresponds to a degenerate solution in which the source generator trivially reduces the path variance by allowing the source samples to diverge.

\section{Direct source--target alignment for complex
training dynamics}
\label{sec:appx_balance}

As discussed in Sec.~\ref{subsec:design}, conditioning the flow model $v_\theta(\cdot)$ with condition $C$ $(\text{i.e., } v_\theta(X_t,t,C))$ can make source learning more challenging from an optimization perspective in complex text-to-image settings. As observed in Appx.~\ref{subsec:appx_complex}, we attribute this to the fact that the source model has less incentive to remain discriminative when conditional information is directly handled by the flow model. 

\begin{figure*}[h]
    \centering
    \includegraphics[width=0.98\linewidth]{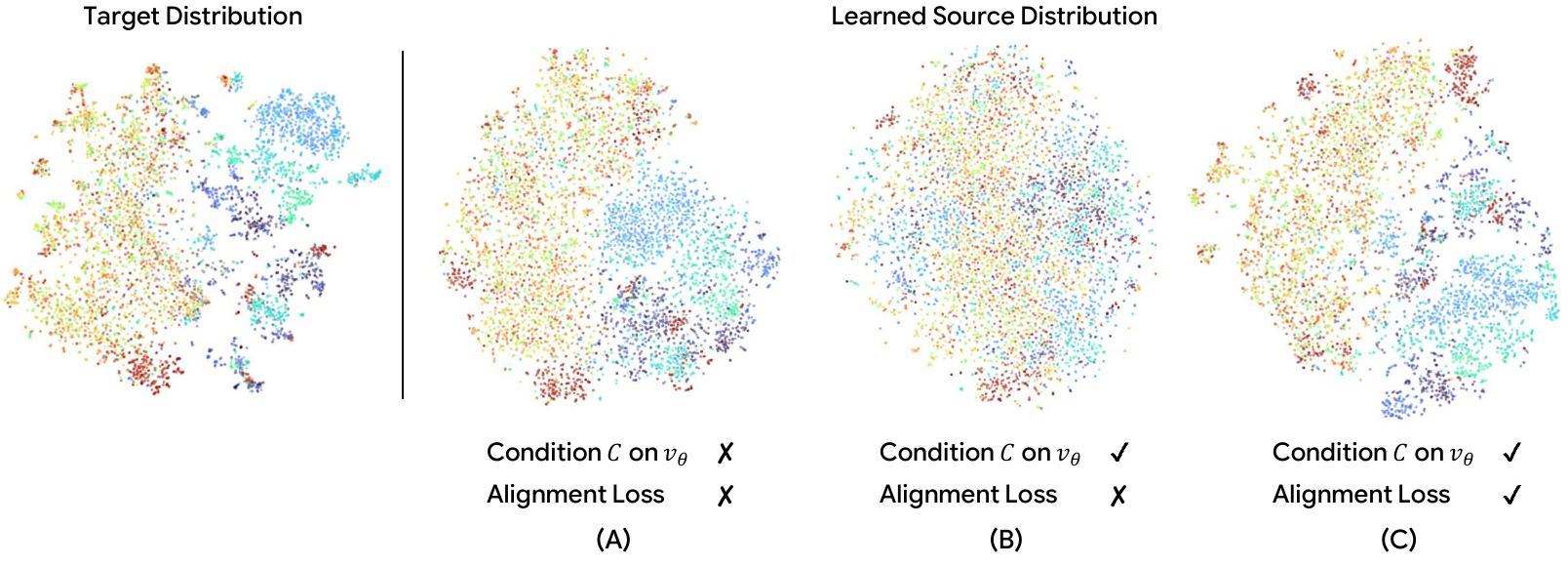}
    \caption{\textbf{t-SNE visualization} of learned source distribution comparing the effects of condition injection and alignment loss.}
    \label{fig:appx_tsne}
\end{figure*}

This effect also manifests in practical settings. We evaluate this behavior using LightningDiT with a learnable source generator and the variance regularization term $\mathcal{L}_{\mathrm{VarReg}}$. As shown in Tab.~\ref{tab:appx_cond}, adding conditioning to the flow model $v_\theta(\cdot)$ degrades FID relative to the unconditioned counterpart, despite improving CLIP score due to stronger conditional modeling. Consistently, Fig.~\ref{fig:appx_tsne} illustrates that conditioning $v_\theta(\cdot)$ causes the learned source distribution to become less discriminative, indicating that the source struggles to learn meaningful structure.

In contrast, when combined with the alignment loss, we observe improvements in both FID and CLIP score, along with a more discriminative and structured source distribution.

\begin{table}[thbp]
\centering
\caption{\textbf{Ablation results analyzing the effects of alignment loss and backbone conditioning.}}
\vspace{0.5em}
\resizebox{0.7\textwidth}{!}{
\begin{tabular}{ccc|ccc}
        \textbf{Flow backbone} & \textbf{Condition $C$ on $v_\theta$} & \textbf{Alignment Loss $\mathcal{L_\mathrm{align}}$}  & \textbf{FID} & \textbf{CLIP}  \\
        \midrule
        \multirow{3}{*}{LightningDiT} & \blackx & \blackx & 3.167 & 0.3335 \\
        & \blackv & \blackx & 3.188 & 0.3354 \\
        & \blackv & \blackv & 3.041 & 0.3363 \\
        \bottomrule
\end{tabular}}
\label{tab:appx_cond}
\vspace{-5pt}
\end{table}

\section{Decomposition of the Flow Matching Objective}
\label{sec:app_fm_decomposition}

We derive the decomposition of the Flow Matching (FM) objective in Eq.~\eqref{eq:fm_loss} into an approximation error term and an intrinsic variance term. Recall that the FM loss is defined as:
\begin{equation}
\mathcal{L}_{\mathrm{FM}}(\theta)
=
\mathbb{E}_{t \sim p(t),\,(X_0,X_1)\sim\pi}
\big[
\| v_\theta(X_t,t) - \Delta \|^2
\big],
\label{eq:app_fm_loss}
\end{equation}
where $\Delta = X_1 - X_0$ and $X_t = (1-t)X_0 + tX_1$. Let
\begin{equation}
u_t(x) := \mathbb{E}[\Delta \mid X_t = x]
\label{eq:app_ut_def}
\end{equation}
denote the marginal velocity field induced by the coupling $\pi$. We rewrite the squared error as:
\begin{align}
\| v_\theta(X_t,t) - \Delta \|^2
&=
\| v_\theta(X_t,t) - u_t(X_t) + u_t(X_t) - \Delta \|^2 .
\end{align}

Expanding the norm yields:
\begin{align}
\| v_\theta(X_t,t) - \Delta \|^2
&=
\| v_\theta(X_t,t) - u_t(X_t) \|^2
+ \| u_t(X_t) - \Delta \|^2
\nonumber \\
&\quad
+ 2 \langle v_\theta(X_t,t) - u_t(X_t),\, u_t(X_t) - \Delta \rangle .
\label{eq:app_expand}
\end{align}

Taking expectation with respect to $t$ and $(X_0,X_1)\sim\pi$, the cross term vanishes:
\begin{align}
&\mathbb{E}\big[
\langle v_\theta(X_t,t) - u_t(X_t),\, u_t(X_t) - \Delta \rangle
\big]
\nonumber \\
&\qquad=
\mathbb{E}\Big[
\mathbb{E}\big[
\langle v_\theta(X_t,t) - u_t(X_t),\, u_t(X_t) - \Delta \rangle
\mid X_t
\big]
\Big]
\nonumber \\
&\qquad=
\mathbb{E}\Big[
\langle v_\theta(X_t,t) - u_t(X_t),\,
\mathbb{E}[u_t(X_t) - \Delta \mid X_t]
\rangle
\Big]
= 0,
\end{align}
since $\mathbb{E}[\Delta \mid X_t] = u_t(X_t)$ by definition. The remaining second term satisfies
\begin{align}
\mathbb{E}\big[
\| u_t(X_t) - \Delta \|^2
\big]
&=
\mathbb{E}\big[
\mathrm{Var}(\Delta \mid X_t)
\big],
\label{eq:app_variance}
\end{align}
which follows directly from the definition of conditional variance. Combining the above results, the FM objective decomposes as:
\begin{align}
\mathcal{L}_{\mathrm{FM}}(\theta)
&=
\mathbb{E}_{t,\pi}
\big[
\| v_\theta(X_t,t) - u_t(X_t) \|^2
\big]
+
\mathbb{E}_{t,\pi}
\big[
\mathrm{Var}(\Delta \mid X_t)
\big].
\label{eq:app_final_decomposition}
\end{align}

The first term corresponds to the approximation error incurred by learning the marginal velocity field $u_t$, while the second term is an intrinsic variance determined solely by the coupling $\pi$, independent of the model parameters $\theta$.

For conditional FM, the same orthogonal decomposition holds after additionally conditioning on $C$. Defining
$u_t(x,c)=\mathbb{E}[\Delta\mid X_t=x,C=c]$, the intrinsic term becomes
\[
\mathbb{E}_{t,\pi(X_0,X_1,C)}[\mathrm{Var}(\Delta\mid X_t,C)].
\]
This term captures the ambiguity that remains when the condition is treated as fully available to the velocity model. In the main text, we present the standard form in Eq.~\eqref{eq:decomposition} because it highlights how the interpolation path changes the information and ambiguity carried by $X_t$ itself.

\section{Training and evaluation details}
\label{sec:appx_train}

We detail the architectures of the source generator in Tab.~\ref{tab:appx_encoder}. The architectures of the flow matching backbones used in Tab.~\ref{tab:abl} and Tab.~\ref{tab:scale_results} are provided in Tab.~\ref{tab:appx_backbone}. For UnifiedNextDiT(1.3B), which is employed in the scaling experiments, we omit the DDT head since its hidden dimensionality is sufficiently large relative to the latent dimension. In the scaling experiments, the model is pretrained for 400K iterations and subsequently finetuned for 15K iterations. Note that due to differences in internal architectural design, models with the same number of layers and hidden dimensionality may still have different total parameter counts.

All models are trained on a TPUv5p cluster with Torch/XLA, with each model taking approximately one day to train. For evaluation, all experiments are conducted on NVIDIA L40S GPUs, and images are generated from validation captions for all reported metrics. 
FID, Inception Score (IS), sFID, and Precision--Recall metrics are computed using ADM precomputed statistics~\cite{dhariwal2021diffusion}. 
CLIP Score~\cite{hessel2021clipscore} is computed using the ViT-B/32 model. 
FDD~\cite{stein2023fdd} is computed using DINOv2 features to provide an additional fidelity-oriented evaluation. 
VQAScore~\cite{lin2024vqascore} is computed with Qwen2.5-VL to further assess text--image alignment.
To balance the scale of source-related objectives with the flow matching loss, we set $\lambda_{\mathrm{VarReg}} = 5.0$ and $\lambda_{\mathrm{align}} = 1.0$ across all experiments.

For the source generator, we take the input text embedding $\mathbf{e}^{\mathrm{text}} \in \mathbb{R}^{N \times d}$, where $N$ denotes the text sequence length and $d$ is the text embedding dimension. We adopt a Perceiver-style architecture~\cite{li2022blip} with $S$ learnable query tokens, where $S$ matches the flattened image-token sequence length. The image embedding $\mathbf{e}^{\mathrm{img}} \in \mathbb{R}^{S \times D}$ serves as the target representation, where $D$ is the image embedding dimension. Since flow matching requires the source and target to lie in the same space, we use cross-attention to map the query tokens conditioned on $\mathbf{e}^{\mathrm{text}}$ into the image embedding space $\mathbb{R}^{S \times D}$.

Following a VAE-style parameterization~\cite{kingma2022autoencodingvariationalbayes}, the source generator uses separate output heads to predict the conditional mean $\mu_\phi(C) \in \mathbb{R}^{S \times D}$ and variance $\sigma_\phi^2(C) \in \mathbb{R}^{S \times D}$. We sample $X_0 \sim \mathcal{N}(\mu_\phi(C), \sigma_\phi^2(C)I)$ using the reparameterization trick. For simplicity and to avoid introducing an additional scale hyperparameter, the variance regularization in Eq.~\eqref{eq:var_reg} uses unit target variance, i.e., target standard deviation $1.0$.

For architectures such as LightningDiT~\cite{yao2025lightning} and MMDiT~\cite{esser2024sd3} that require AdaLN-based modulation, we use the pooled text representation when available; otherwise, we fall back to a mean-pooled text embedding.

At inference, the source generator is evaluated only once to sample the initial source $X_0$. For $256\times256$ image generation with 50 NFE, the default sampling setting used throughout the paper, this adds only $0.36$ ms of average latency, which is negligible compared with the $91.83$ ms average ODE sampling time of the velocity model.

\begin{table}[thbp]
\centering
\caption{\textbf{Architectural details of the source generator used in our experiments.}}
\vspace{0.5em}
\label{tab:appx_encoder}
\begin{tabular}{lcc}
\toprule
    & \multicolumn{2}{c}{\textbf{Source Generator}}  \\
\midrule
\textbf{Architecture} & \\
Input dim. & 768 & 1024 \\
Hidden dim. & 768 & 768 \\
Num. layers & 8 & 8 \\
Num. heads & 12 & 12 \\
\midrule
Params & 77M & 78M \\
Text Encoder & CLIP & Qwen3 \\
Max length & 77 & 128 \\
\bottomrule
\end{tabular}
\end{table}

\begin{table}[thbp]
\centering
\caption{\textbf{Architectural and optimization details of flow matching backbones used in our experiments.}}
\label{tab:appx_backbone}
\resizebox{\textwidth}{!}{
\begin{tabular}{lccccc}
\toprule
 & \textbf{LightningDiT} & \textbf{MMDiT} & \textbf{UnifiedNextDiT} & 
 \textbf{UnifiedNextDiT$^\dagger$} & \textbf{UnifiedNextDiT(1.3B)} \\
\midrule
\textbf{Architecture} &  &  \\
Downsample Ratio & $16 \times 16$ & $16 \times 16$ & $16 \times 16$  & $16 \times 16$ & $14 \times 14$ \\
Latent dim. & 768 & 768 & 768 & 768 & 1152 \\
Num. layers & 12 & 12 & 14 & 16 & 20 \\
Hidden dim. & 768 & 768 & 864 & 864 & 1728 \\
Num. heads & 12 & 12 & 12 & 12 & 16\\
\midrule
\textbf{DDT Head} &  &  \\
Num. layers & 2 & 2 & 2 & 2 & \multirow{3}{*}{\blackx}\\
Hidden dim. & 2048 & 2048 & 2048 & 2048 \\
Num. heads & 16 & 16 & 16 & 16 \\
\midrule
Params & 294M & 418M & 414M & 471M & 1.24B\\
Base Encoder & DINOv2 & DINOv2 & DINOv2 & DINOv2 & SigLIP2 \\
\midrule
\textbf{Optimization} &  &  \\
Batch Size & 1024 & 1024 & 1024 & 1024 & 256 \\
Optimizer & AdamW & AdamW & AdamW & AdamW & AdamW \\
Base lr & $2 \times 10^{-4}$ & $2 \times 10^{-4}$ & $2 \times 10^{-4}$ & $2 \times 10^{-4}$ & $2 \times 10^{-4}$  \\
$(\beta_1, \beta_2)$ & (0.9, 0.95) & (0.9, 0.95) & (0.9, 0.95) & (0.9, 0.95) & (0.9, 0.95) \\
\bottomrule
\end{tabular}
}
\end{table}

\section{Comparison implementation and interpretation}
\label{sec:appx_impl_detail_comparison}

We provide implementation details for the comparison methods reported in Tab.~\ref{tab:learned_source_comparison}. For each method, we follow the official implementation as closely as possible while adapting it to the shared RAE (DINOv2) target representation and UnifiedNextDiT~\cite{qin2025lumina} backbone with a DDT head~\cite{wang2025ddt}. All models are trained for 100K steps with batch size 1024 and learning rate $2\times 10^{-4}$, using CLIP~\cite{radford2021learning} as the text encoder. For C$^2$OT~\cite{cheng2025c2ot}, we use an oversampling ratio of 10 for the OT coupling batch. For CPD~\cite{issachar2025designing}, we pretrain the source generator for 100K steps before training the flow model. For CrossFlow, CPD, and CAR-Flow, we match the number of parameters used in the condition-dependent source generator.

The results in Tab.~\ref{tab:learned_source_comparison} suggest that simply introducing condition information into the source is not sufficient for robust gains in this high-dimensional T2I setting. CPD constructs the source through a separate regression objective before FM training; while this improves source--target proximity, the source is not optimized with the velocity model under the FM objective. CAR-Flow jointly optimizes a condition-aware reparameterization, but its constrained source parameterization provides limited flexibility in adapting the stochastic support of the source. C$^2$OT directly modifies sample-level coupling, but reliable minibatch matching becomes difficult at this scale. CrossFlow demonstrates that text-dependent sources are feasible, but its cross-modal transport formulation does not fully exploit the strongly conditioned T2I backbone used here. In contrast, \ours{} learns the source under the FM objective while explicitly stabilizing source variance and strengthening source supervision, leading to the strongest performance across fidelity and alignment metrics.

\section{Guidance}
\label{sec:appx_guidance}

Because the learned source $X_0 \sim p_\phi(X_0\mid C)$ already encodes conditional information, \ours{} does not directly provide a separate unconditional branch for standard classifier-free guidance (CFG)~\cite{ho2022cfg}. Although CFG could be introduced through additional dropped- or mismatched-condition training, as in CrossFlow~\cite{liu2025crossflow}, we leave this extension unexplored since it is orthogonal to our focus on condition-dependent source learning.

We instead adopt AutoGuidance~\cite{karras2024ag}, following prior observations that it is effective in RAE feature spaces~\cite{zheng2025rae}. We select guidance scales by grid search under the same evaluation protocol. As shown in Tab.~\ref{tab:guidance}, AutoGuidance improves both the Gaussian-source FM baseline and \ours{}, indicating that guidance remains effective with learned condition-dependent sources. For reference, we also report a CFG-guided Gaussian-source FM baseline: CFG improves unguided FM, but remains worse than unguided \ours{} in FID. Its higher CLIP Score should be interpreted carefully, since it exceeds the validation-caption CLIP Score of $0.3452$ and may reflect over-optimization toward the CLIP objective rather than uniformly better generation quality.

\begin{table}[thbp]
\centering
\vspace{-1em}
\caption{\textbf{Guidance analysis on ImageNet 256$\times$256.} We evaluate whether guidance remains effective with learned condition-dependent sources by applying AutoGuidance (AG)~\cite{karras2024ag} to both the Gaussian-source FM baseline and \ours. AG improves both methods, while \ours{} with AG achieves the best FID and maintains a strong CLIP Score. For reference, we also report CFG on the Gaussian-source baseline.}
\vspace{0.5em}
\resizebox{0.4\textwidth}{!}{
\begin{tabular}{l |ccc}
        \textbf{Method} & \textbf{FID~$\downarrow$} & \textbf{CLIP~$\uparrow$}  \\
        \midrule
        \midrule
        Standard FM & 3.036 & 0.3398 \\
        Standard FM $+$ CFG & 2.801 & \textbf{0.3463} \\ 
        Standard FM $+$ AG & 2.673 & 0.3400 \\
        \midrule
        CSFM (Ours) & 2.453 & 0.3420 \\
        \rowcolor{blue!10} \textbf{CSFM (Ours) $+$ AG} & \textbf{2.016} & 0.3433 \\
        \bottomrule
\end{tabular}}
\label{tab:guidance}
\end{table}

\section{ImageNet captioned dataset}
\begin{figure*}[h]
    \centering
    \includegraphics[width=0.98\linewidth]{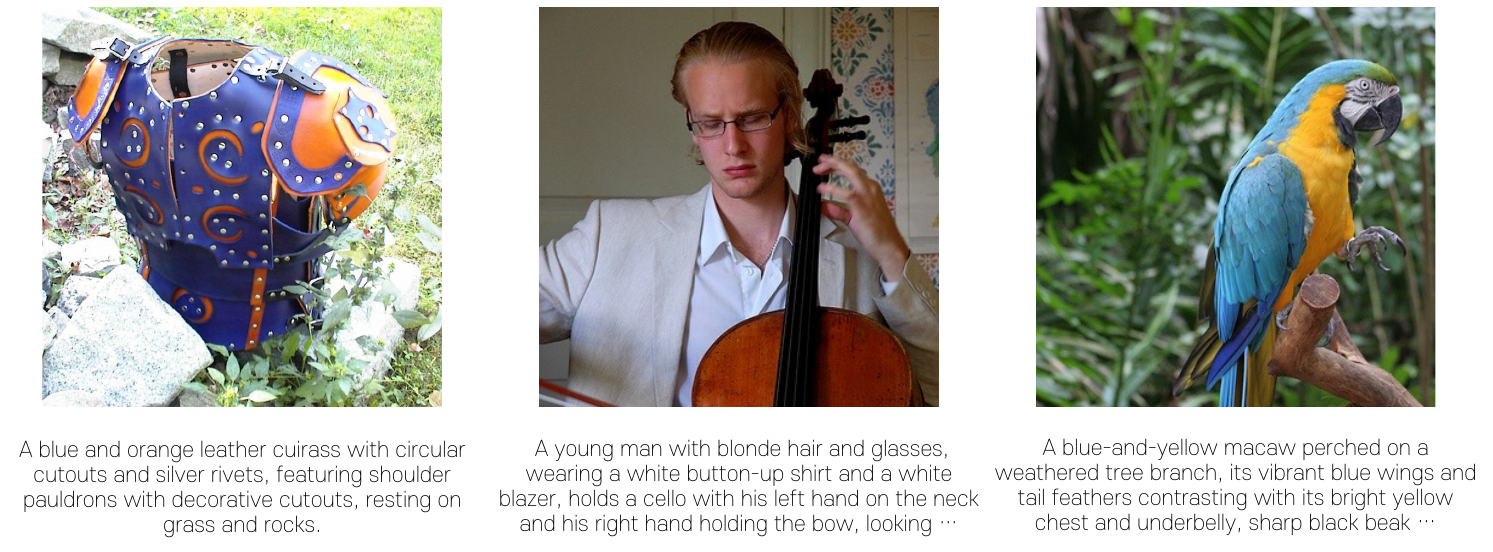}
    \caption{\textbf{Examples of detailed image–caption pairs from the constructed captioned ImageNet-1K dataset.}}
    \label{fig:appx_dataset}
\end{figure*}

As modern text-to-image frameworks increasingly emphasize scaling, there is no well-established setting for component-wise analysis~\cite{degeorge2025far}. To enable controlled component-wise experiments at a manageable scale, we construct a captioned dataset based on ImageNet-1K~\cite{russakovsky2015imagenet}. Detailed image captions are generated using Qwen3-VL-8B-Instruct~\cite{Qwen3-VL}. Examples from the resulting dataset are shown in Fig.~\ref{fig:appx_dataset}.
\section{Gradient variance}
\label{subsec:appx_gradvar}
While we can directly estimate intrinsic variance in the low-dimensional toy setting (Appx.~\ref{sec:appx_toy}), doing so in high-dimensional continuous latent spaces is difficult, since it would require reliable samples with the same, or sufficiently nearby, intermediate state $X_t$. Such estimates are highly sensitive to neighborhood choice and become unreliable as dimensionality increases. We therefore use gradient variance as a practical diagnostic of supervision ambiguity induced by the source--target coupling.

We report the variance of gradients of the flow model $v_\theta(\cdot)$ induced by the Flow Matching loss, i.e., $\mathrm{Var}(\nabla_\theta \mathcal{L}_{\mathrm{FM}})$, across the interpolation time $t$ (Fig.~\ref{fig:fmloss_gradvar}). 
To compute gradients, we select five attention layers uniformly sampled from the flow backbone and measure the gradients of the MLP parameters that project inputs to the query, key, and value vectors.
The reported variance is further averaged across the selected layers.
Due to the small magnitude of individual gradient components, we compute the variance element-wise and report the sum across dimensions.
Experiments are conducted using 20K samples from the ImageNet-1K~\cite{russakovsky2015imagenet} training set and a model trained for 100K steps.
We use \texttt{torch.func.vmap} to efficiently compute per-sample gradients.


\section{Detailed related work}
\label{sec:appx_relwork}

\paragraph{Flow Matching.}
Flow Matching (FM) learns a velocity field that defines a continuous probability path between a source distribution and a target distribution~\cite{lipman2022fm, liu2022rf, lipman2024flowmatchingguidecode}. Owing to its conceptual simplicity and strong generative performance, FM has become a widely used framework for generative modeling. Unlike conventional diffusion formulations~\cite{ho2020ddpm}, FM does not restrict the source distribution to a standard Gaussian~\cite{liu2025crossflow}. Despite this flexibility, most practical systems still adopt a fixed Gaussian source, leaving the design and learning of source distributions comparatively underexplored.

\paragraph{Optimal Transport coupling.}
A standard independent coupling between Gaussian source samples and target data can induce intersecting training trajectories, leading to curved flows and high-variance FM gradients. To mitigate this issue, OT-based methods couple source and target samples using minibatch-level optimal transport plans~\cite{tong2023otcfm, ot-multisample, kong2025alignflowimprovingflowbasedgenerative}. C$^2$OT~\cite{cheng2025c2ot} further analyzes OT coupling in conditional generation and proposes condition-aware OT to address conditionally skewed source distributions. While these methods clarify the importance of source--target coupling, they typically require costly minibatch optimization and become difficult to estimate reliably in high-dimensional conditional settings. In contrast, our method learns condition-dependent source--target couplings end-to-end through the FM objective.

\paragraph{Condition-dependent source distributions.}
Several recent works investigate condition-dependent or structured source distributions for conditional flow matching. CPD~\cite{issachar2025designing} constructs a condition-dependent prior through a separate regression objective before flow model training, while \citet{ahn2024noiseworthdiffusionguidance} learn a source generator from diffusion-inversion trajectories. Other works study source design from geometric, optimization, or reparameterization perspectives~\cite{lee2025there, chen2025carflow}. Broader conditional FM frameworks, including WFM~\cite{haviv2024wfm}, Meta Flow Matching~\cite{atanackovic2024meta}, and Cocos~\cite{dong2025cocos}, address related but distinct problems such as conditional parameterization, task adaptation, or distribution-level transport. In contrast, we study jointly learned condition-dependent sources as a mechanism for reducing intrinsic variance and improving FM dynamics in modern T2I systems.

\paragraph{Flow Matching between distributions.}
Another line of work studies flow matching between different distributions or modalities. Early bridge-based methods~\cite{shi2023diffusion, tong2023simulationfree, zhou2023denoising} primarily focus on related domains such as image-to-image translation. CrossFlow~\cite{liu2025crossflow} formulates text-to-image generation as cross-modal flow matching, and FlowTok~\cite{he2025flowtok} extends this direction to tokenized one-dimensional representations. VAFlow~\cite{Wang_2025_ICCV} similarly applies cross-modal flow matching to video-to-audio generation. These approaches demonstrate that flow matching can transport across modalities, but their main goal is cross-modal distribution-to-distribution transport. In contrast, we study condition-dependent source learning as a design problem for improving FM dynamics themselves, connecting it to intrinsic variance, stable optimization, and representation-dependent regimes of effectiveness.

\section{Prompt-conditional diversity.}

To verify that the improved fidelity does not come from a substantial loss of diversity, we measure prompt-conditional multimodality~\cite{zhang2023t2mgpt}. For each prompt, we generate 20 images, form 10 random pairs, compute the average pairwise distance in feature space, repeat this five times per prompt, and average over 500 validation prompts. \ours{} shows only a modest decrease in multimodality compared to FM, while substantially improving FID. We also report FM with CFG as a reference point, since CFG is known to trade diversity for fidelity.

\begin{table}[h]
\centering
\vspace{-1em}
\caption{\textbf{Prompt-conditional diversity analysis on ImageNet 256$\times$256.} We evaluate whether the fidelity improvement of \ours{} comes at the cost of reduced sample diversity. We report multimodality (MModality), computed from pairwise feature distances among multiple samples generated from the same prompt.}
\vspace{0.5em}
\resizebox{0.4\textwidth}{!}{
\begin{tabular}{l |cc}
        \textbf{Method} & \textbf{FID~$\downarrow$} & \textbf{MModality~$\uparrow$}  \\
        \midrule
        \midrule
        Standard FM & 3.036 & 9.6224 $\pm$ 0.3837 \\
        Standard FM $+$ CFG & 2.801 & 8.2333 $\pm$ 0.3300 \\
        \midrule
        \rowcolor{blue!10} \textbf{CSFM (Ours)} & \textbf{2.453} & 9.2367 $\pm$ 0.3702 \\
        \bottomrule
\end{tabular}}
\label{tab:diversity}
\end{table}

\section{Additional target representation analysis}
\label{sec:appx_rep_other_method}

\begin{figure*}[h]
    \centering
    \includegraphics[width=0.48\linewidth]{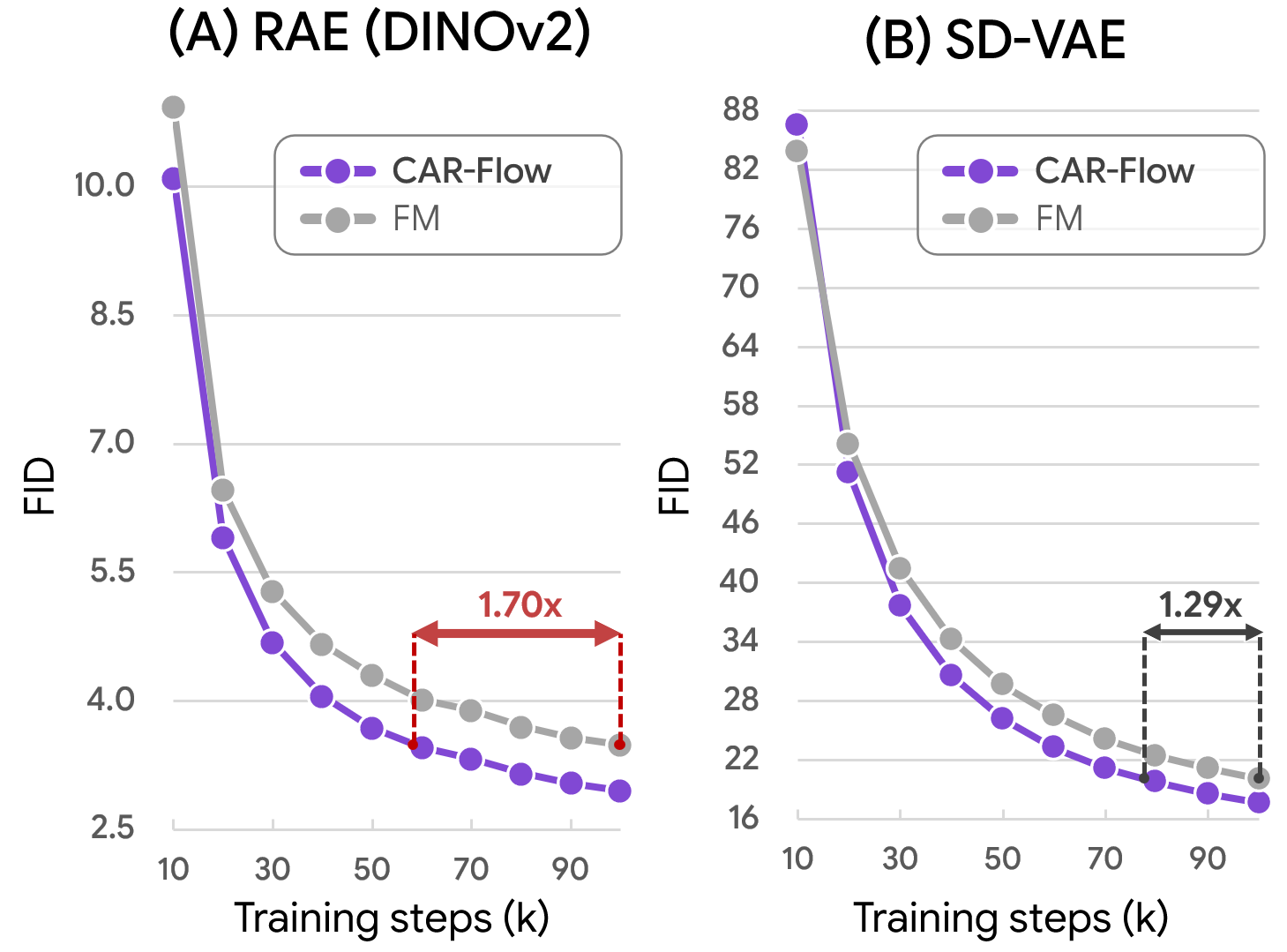}
    \caption{\textbf{Target representation analysis with CAR-Flow.} We compare CAR-Flow in SD-VAE and RAE (DINOv2) representation spaces. The larger gain in RAE space indicates that the importance of structured target representations generalizes beyond \ours{}.}
    \label{fig:carflow_rae_sdvae}
\end{figure*}

We further evaluate whether the effect of target representation generalizes beyond \ours{} by repeating the RAE-versus-SD-VAE comparison with CAR-Flow~\cite{chen2025carflow}. We use the same implementation setting as in Appx.~\ref{sec:appx_impl_detail_comparison}, except that these experiments are conducted with batch size 512. As shown in Fig.~\ref{fig:carflow_rae_sdvae}, CAR-Flow also obtains larger gains in the RAE (DINOv2) representation space than in the SD-VAE space. This supports our conclusion that the target representation geometry is an important factor for condition-dependent source design, beyond the specific choice of \ours{}.

\section{Additional quantitative results}
\label{sec:appx_full_benchmark}
Tab.~\ref{tab:appx_geneval} reports the full benchmark comparison on GenEval~\cite{ghosh2023geneval}, and Tab.~\ref{tab:appx_dpg} reports the full benchmark comparison on DPG-Bench~\cite{hu2024ella} between standard FM and CSFM.

\begin{table}[h]
\centering
\caption{\textbf{Category-wise performance comparison of standard FM and CSFM on GenEval.}}
\label{tab:appx_geneval}
\vspace{0.5em}
\resizebox{0.9\textwidth}{!}{%
\begin{tabular}{l|ccccccc}
\toprule
Model & Single Obj. & Two Obj. & Counting & Colors & Position & Color Attri. & Overall$\uparrow$ \\
\midrule
Standard FM & 0.99 & 0.87 & 0.63 & 0.85  & 0.76  & 0.53 & 0.77 \\
CSFM (Ours)& 0.98 & 0.88 & 0.65 & 0.90 & 0.79 & 0.59 & 0.80 \\
\bottomrule
\end{tabular}%
}
\end{table}

\begin{table}[h]
\centering
\caption{\textbf{Category-wise performance comparison of standard FM and CSFM on DPG-Bench.}}
\label{tab:appx_dpg}
\vspace{0.5em}
\begin{tabular}{l|cccccc}
\toprule
Model & Global & Entity & Attribute & Relation & Other & Overall$\uparrow$ \\
\midrule
Standard FM & 81.46 & 87.04 & 85.06 & 92.80 & 75.60 & 78.31 \\
CSFM (Ours) & 84.80 & 88.80 & 85.93 & 93.23 & 76.00 & 81.11 \\
\bottomrule
\end{tabular}
\end{table}

\section{Additional qualitative results}
\begin{figure}[h]
    \centering
    \includegraphics[width=0.98\linewidth]{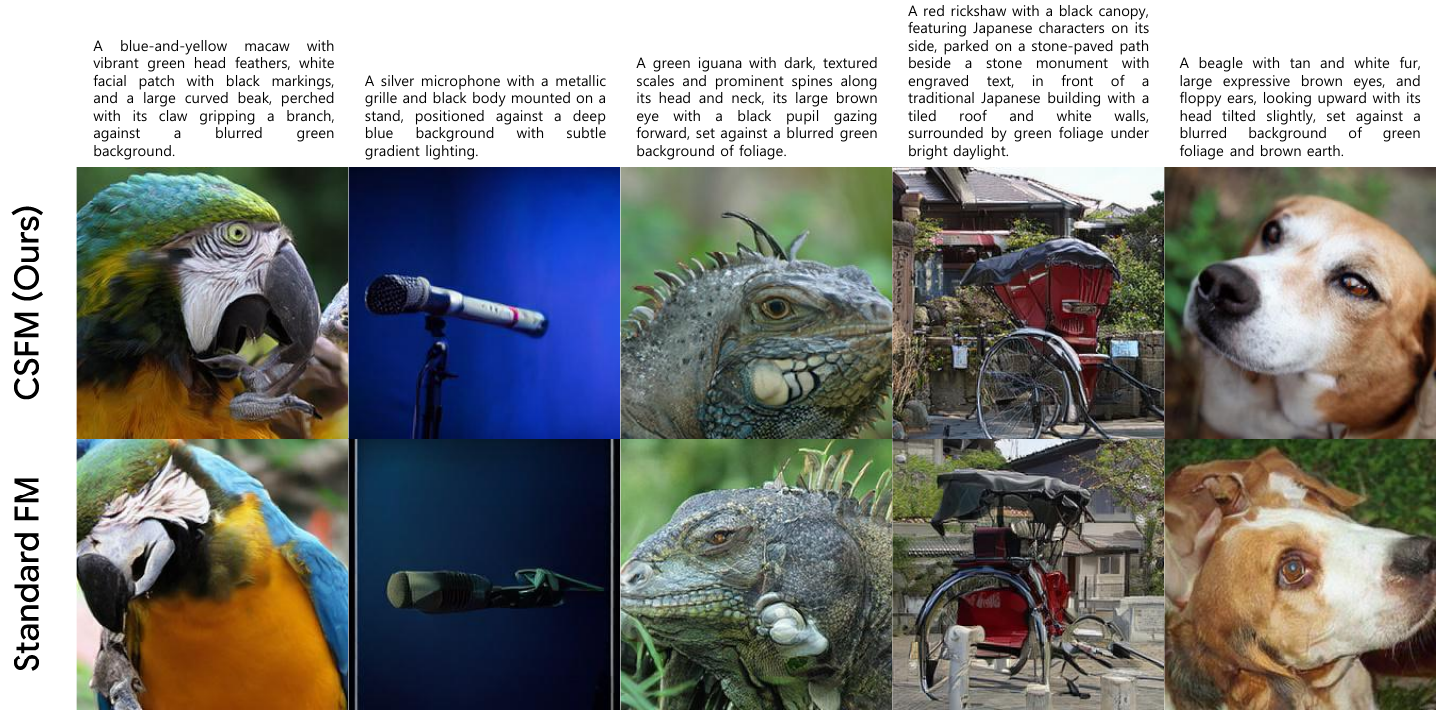}
    \caption{Qualitative comparison between fixed and learned source distributions on ImageNet-1K~\cite{russakovsky2015imagenet}.}
    \label{fig:appx_imagenet_qual1}
\end{figure}

\begin{figure}[h]
    \centering
    \includegraphics[width=0.98\linewidth]{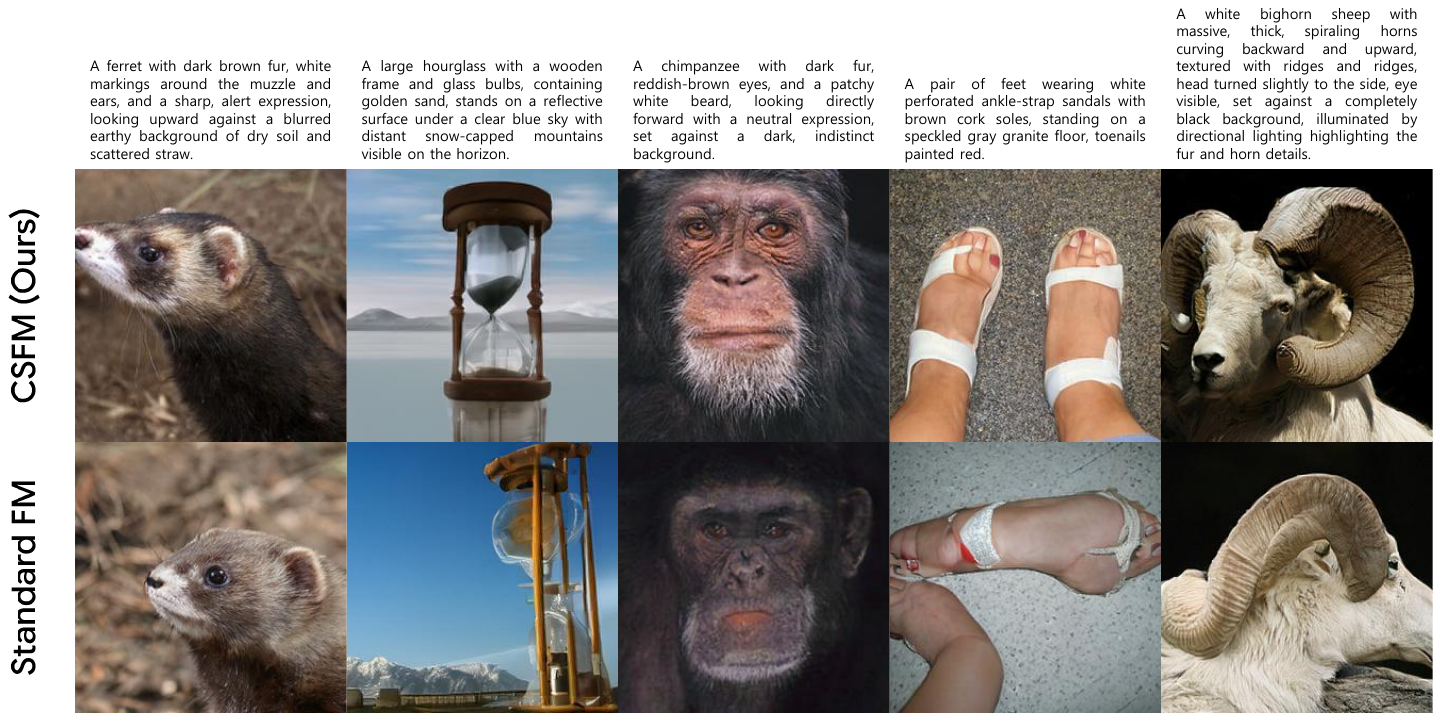}
    \caption{Qualitative comparison between fixed and learned source distributions on ImageNet-1K~\cite{russakovsky2015imagenet}.}
    \label{fig:appx_imagenet_qual2}
\end{figure}

\begin{figure*}[h]
    \centering
    \includegraphics[width=0.98\linewidth]{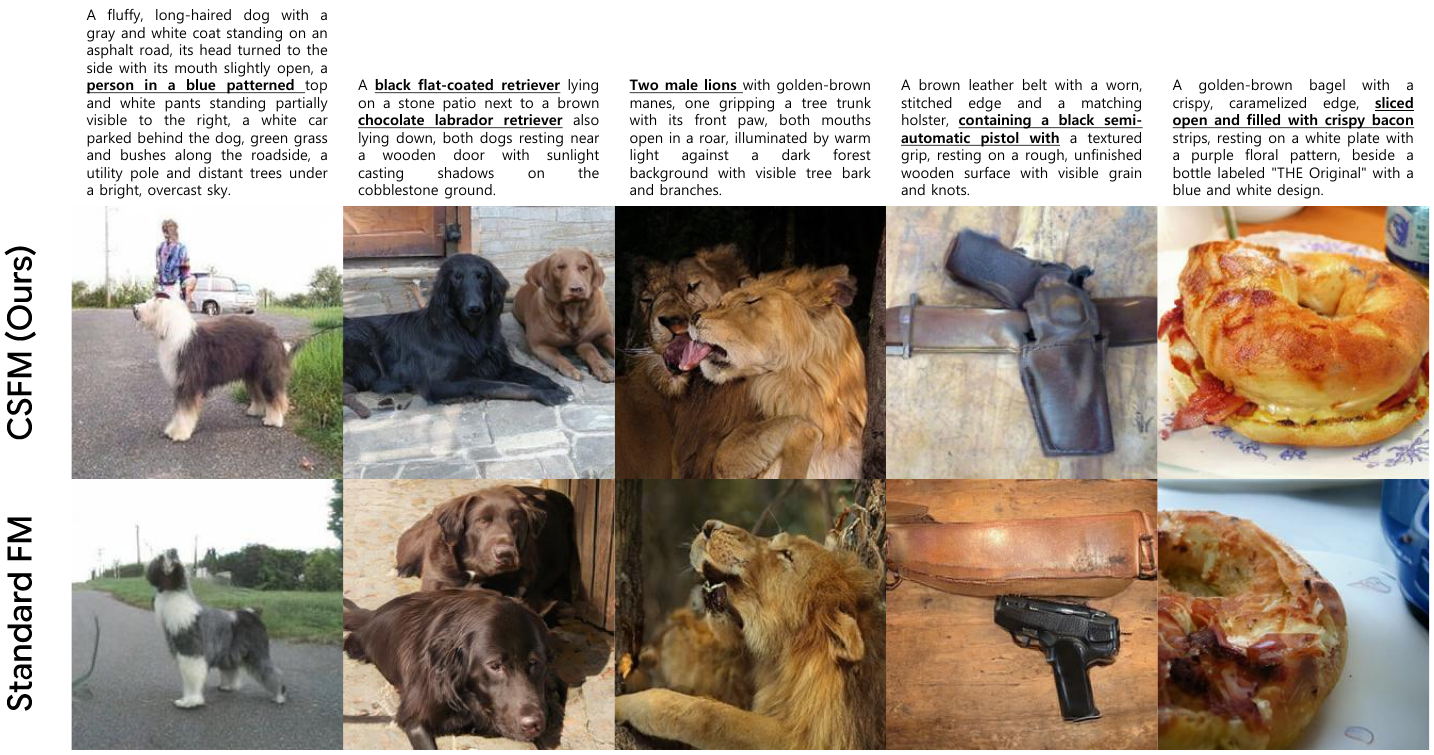}
    \caption{Qualitative comparison between fixed and learned source distributions for prompts with \underline{\textbf{multiple objects}}  and \underline{\textbf{complex relationships}}, on ImageNet-1K~\cite{russakovsky2015imagenet}.}
    \label{fig:appx_imagenet_qual_complexcaption}
\end{figure*}

\begin{figure}[h]
    \centering
    \includegraphics[width=0.98\linewidth]{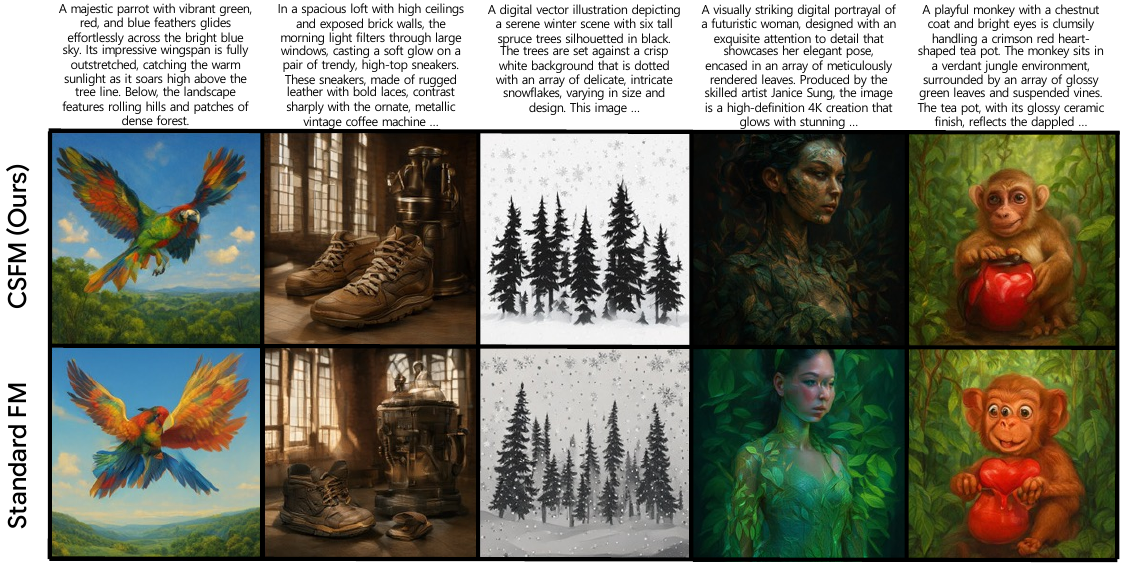}
    \caption{Qualitative comparison between fixed and learned source distributions using UnifiedNextDiT (1.3B).}
    \label{fig:appx_qual1}
\end{figure}

\begin{figure}[h]
    \centering
    \includegraphics[width=0.98\linewidth]{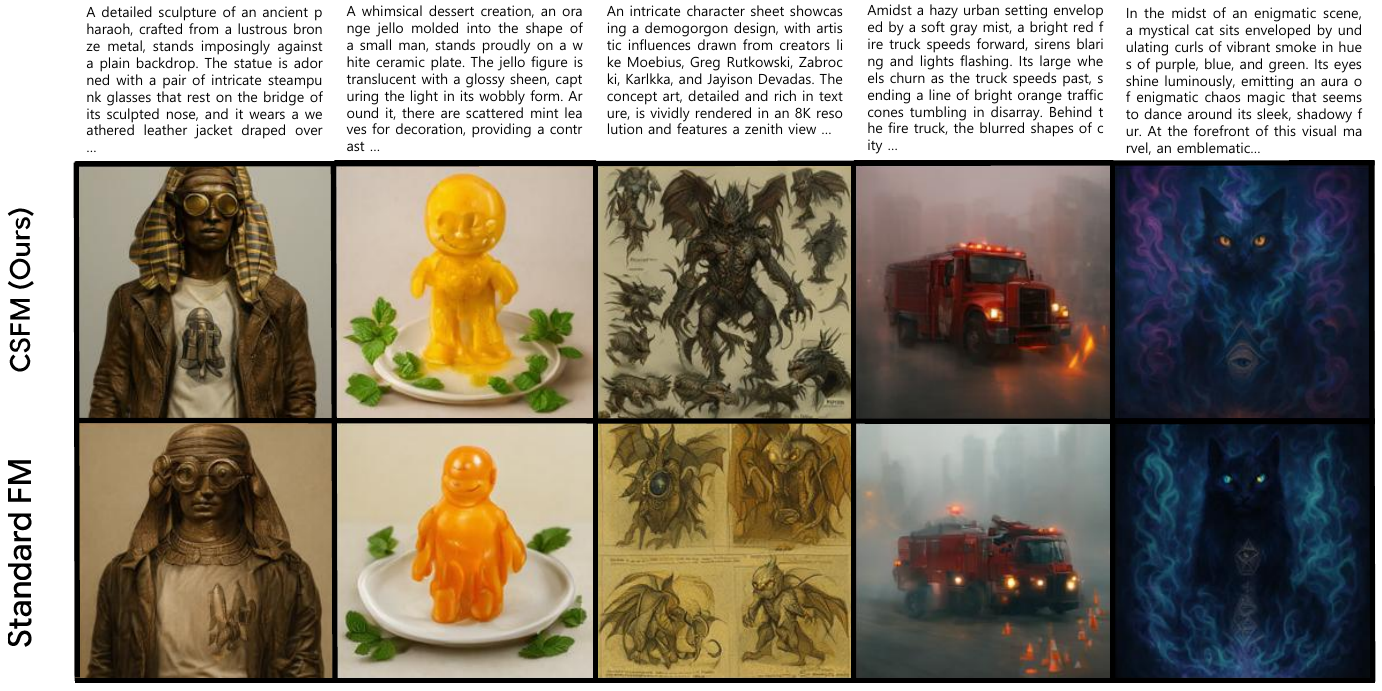}
    \caption{Qualitative comparison between fixed and learned source distributions using UnifiedNextDiT (1.3B).}
    \label{fig:appx_qual2}
\end{figure}

\begin{figure}[h]
    \centering
    \includegraphics[width=0.98\linewidth]{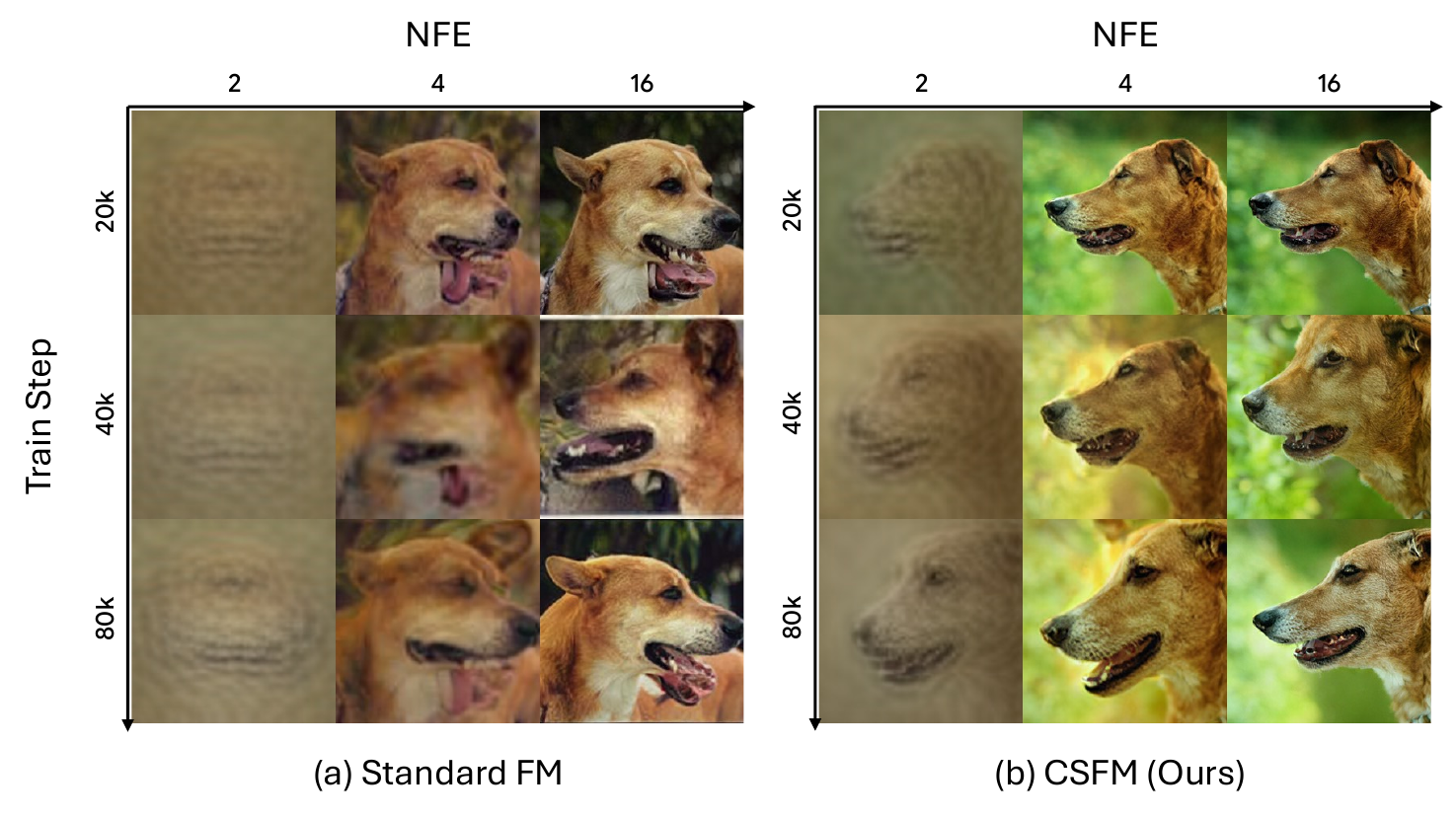}
    \caption{Qualitative comparison between standard FM and CSFM across training steps and NFEs (Number of Function Evaluations) on ImageNet-1K~\cite{russakovsky2015imagenet}.}
    \label{fig:appx_imagenet_qual_trainstep_nfe}
\end{figure}

We present qualitative results comparing fixed and learnable source distributions.
Fig.~\ref{fig:appx_imagenet_qual1} and Fig.~\ref{fig:appx_imagenet_qual2} present qualitative comparisons using the model trained for 100K steps in Sec.~\ref{subsec:main_result}, evaluated on ImageNet-1K~\cite{russakovsky2015imagenet} validation set, with a 50-step Euler ODE sampler. Fig.~\ref{fig:appx_imagenet_qual_complexcaption} further demonstrates model performance on text involving \textbf{complex relationships} and \textbf{multiple objects}, using the same 50-step Euler ODE sampler.
Fig.~\ref{fig:appx_qual1} and Fig.~\ref{fig:appx_qual2} present qualitative comparisons using the scaled UnifiedNextDiT (1.3B). Fig.~\ref{fig:appx_imagenet_qual_trainstep_nfe} presents qualitative comparison across the training steps and the NFEs (Number of Function Evaluations), demonstrating faster convergence and straighter flow for CSFM. 

\section{Limitations}
\label{sec:appx_limitations}

Our study uses text-to-image generation as a high-dimensional, large-scale testbed for condition-dependent source learning. Although we evaluate across multiple benchmarks, architectures, and a 1.3B-parameter setting using publicly accessible data and compute, further validation at proprietary frontier-model scale remains future work. We also focus on AutoGuidance rather than standard classifier-free guidance, since \ours{} does not naturally include an unconditional branch and AutoGuidance is well suited to RAE feature spaces. Finally, while text-to-image generation provides a representative setting for complex conditional generation, the behavior of CSFM in other domains, such as text-to-audio or text-to-video generation, remains an open direction.

\section{Impact Statement}
\label{sec:appx_impact_statement}
This paper presents work whose goal is to advance the field of Machine Learning, specifically in the area of generative modeling. While generative models have many potential societal consequences, these are largely well understood in the existing literature. As with other generative image models, the proposed method may generate harmful or inappropriate content depending on the training data and prompts, highlighting the importance of responsible dataset curation and deployment. We do not identify additional societal impacts that require specific discussion beyond these established considerations.


\clearpage

\end{document}